\pdfoutput=1

\documentclass{article}
\usepackage{acl}
\usepackage{xcolor}

\usepackage{times}
\usepackage{latexsym}

\usepackage{microtype}
\usepackage{hyperref}
\usepackage{url}
\usepackage{booktabs}
\usepackage{lineno}
\usepackage[T1]{fontenc}

\usepackage[utf8]{inputenc}

\usepackage{inconsolata}

\usepackage{amsmath,amssymb,amsfonts}
\usepackage{mathtools, cuted}
\usepackage{graphicx}
\usepackage{textcomp}
\usepackage{cleveref}
\usepackage{glossaries}
\usepackage{hyperref}
\usepackage{algorithm}
\usepackage{algpseudocode}
\usepackage{xspace,mfirstuc,tabulary}
\usepackage{multirow}
\usepackage{todonotes}
\usepackage{balance}
\usepackage{enumitem}
\usepackage{listings}
\usepackage{subfig}
\usepackage[subfigure]{tocloft}
\usepackage{dblfloatfix}

\definecolor{codegreen}{rgb}{0,0.6,0}
\definecolor{codegray}{rgb}{0.5,0.5,0.5}
\definecolor{codepurple}{rgb}{0.58,0,0.82}
\definecolor{genbackcolour}{rgb}{0.92,0.95,0.92}
\definecolor{backcolour}{rgb}{0.97,0.97,0.97}

\lstdefinestyle{mystyle}{
  backgroundcolor=\color{backcolour}, commentstyle=\color{codegreen},
  keywordstyle=\color{magenta},
  numberstyle=\tiny\color{codegray},
  stringstyle=\color{codepurple},
  emphstyle={\color{cyan}},
  basicstyle=\fontsize{7}{7.8}\selectfont\ttfamily,
  breakatwhitespace=false,         
  breaklines=true,                 
  captionpos=b,                    
  keepspaces=true,                 
  numbersep=5pt,                  
  showspaces=false,                
  showstringspaces=false,
  showtabs=false,                  
  tabsize=2,
  breakindent=1ex
}

\lstdefinestyle{generated}{
  backgroundcolor=\color{genbackcolour}, commentstyle=\color{codegreen},
  keywordstyle=\color{magenta},
  numberstyle=\tiny\color{codegray},
  stringstyle=\color{codepurple},
  emphstyle={\color{cyan}},
  basicstyle=\fontsize{7}{7.8}\selectfont\ttfamily,
  breakatwhitespace=false,         
  breaklines=true,                 
  captionpos=b,                    
  keepspaces=true,                 
  numbersep=5pt,                  
  showspaces=false,                
  showstringspaces=false,
  showtabs=false,                  
  tabsize=2,
  breakindent=1ex
}

\lstdefinelanguage{prompt}
{
  alsoletter=-:,
  morekeywords={
    Instructions:,
    Context:,
    Parents:,
    Reasoning:,
    Interpretation:,
    Child:,
    Parent:,
    Description:,
    Candidates:,
    Leaf:,
    Children:,
    Previous ,
    Taxonomy ,
  },
  emph={---},
  sensitive=true, 
  morecomment=[l]{//}, 
  morestring=*[d]{```} 
}

\algnewcommand\algorithmicforeach{\textbf{for each:}}
\algnewcommand\ForEach{\item[ \algorithmicforeach]}

\makeatletter
\newcommand*{\glsplainhyperlink}[2]{%
  \colorlet{currenttext}{.}
  \colorlet{currentlink}{\@linkcolor}
  \hypersetup{linkcolor=currenttext}
  \hyperlink{#1}{#2}%
  \hypersetup{linkcolor=currentlink}
}
\let\@glslink\glsplainhyperlink
\makeatother
\glsdisablehyper

\DeclareMathAlphabet\mathbfcal{OMS}{cmsy}{b}{n}

\title{FoodTaxo: Generating Food Taxonomies with Large Language Models}

\date{August 2023}

\newacronym{LLM}{LLM}{large language model}
\newacronym{NLIV}{NLIV}{NLI-verification}
\newacronym{NLP}{NLP}{natural language processing}
\newacronym{RAG}{RAG}{retrieval augmented generation}
\newacronym{DSP}{DSP}{demonstrate-search-predict}
\newacronym{CoT}{CoT}{chain-of-thought}
\newacronym{NLI}{NLI}{natural language inference}
\newacronym{WPS}{WPS}{Wu \& Palmer similarity}
\newacronym{CSC}{CSC}{concept similarity correlation}
\newacronym{GRU}{GRU}{gated recurrent unit}

\author{
  Pascal Wullschleger$^{\diamond,\dagger}$,
  Majid Zarharan$^{\diamond}$,
  Donnacha Daly$^\dagger$\\
  \bf{Marc Pouly$^{\dagger}$,
  Jennifer Foster$^{\diamond}$} \\\\
  $^\diamond$ ADAPT Centre, School of Computing, Dublin City University
  \\
  $^\dagger$
  Lucerne School of Computer Science and Information Technology (HSLU)
  \\
  \texttt{pascal.wullschleger@hslu.ch}
}

\newfloat{lstfloat}{htbp}{lop}
\floatname{lstfloat}{Listing}

\begin{document}

{\makeatletter\acl@finalcopytrue
  \maketitle
}

\begin{abstract}
We investigate the utility of Large Language Models for automated taxonomy generation and completion specifically applied to taxonomies from the food technology industry. We explore the extent to which taxonomies can be completed from a seed taxonomy or generated without a seed from a set of known concepts, in an iterative fashion using recent prompting techniques.
Experiments on five taxonomies using an open-source LLM (Llama-3), while promising, point to the difficulty of correctly placing inner nodes. 
\end{abstract}

\section{Introduction}

In the food technology industry, taxonomies play a crucial role in business processes related to generation of new consumer and industrial recipes and the adaption thereof to new culinary trends, diets, and sustainability goals. By replacing ingredients in recipes, one can accommodate allergies and dietary restrictions, lower the carbon footprint, react to supply-chain issues, respect seasonality and avoid food waste. The replacement process can, however, be very complex. Veganizing a dessert or cake recipe by replacing eggs influences the entire cooking process. Likewise, changing the type of nuts in a convenience food recipe can have far-reaching consequences for the whole production line, e.g. due to a different fat percentage. 


To address these challenges, we investigate the automated generation and completion of taxonomies, i.e.~learning taxonomies from data, adding new concepts to existing taxonomies with no human involvement, thereby scaling taxonomies beyond what can be managed by human experts. 




\begin{figure}[t]
\centering\includegraphics[width=1\linewidth]{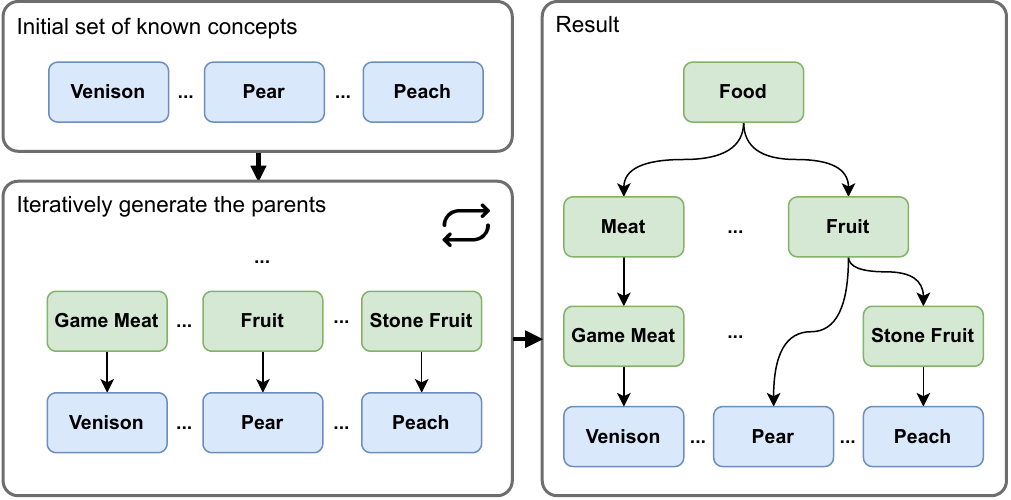}
\caption{The basic intuition behind the generation process. We start with a set of known concepts and iteratively construct a taxonomy in a bottom-up procedure by prompting \glspl{LLM}.}
\label{fig:intro}
\end{figure}

Classical taxonomy completion typically involves extracting concepts from a corpus. However, we suggest that it is often more practical to start with a set of known concepts and extend the set while establishing taxonomic relationships. We hypothesize that taxonomies can be iteratively generated using \glspl{LLM}, without the need for traditional concept extraction (see Fig.~\ref{fig:intro}). This is supported by the state-of-the-art performance of in-context learning with \glspl{LLM} across a range of \gls{NLP} tasks, even without the need for fine-tuning, e.g., \cite{pmlr-v202-zhang23m, milios-etal-2023-context}. Such an approach is particularly advantageous in situations where
it is challenging to provide a suitable corpus for concept extraction. 

We evaluate our proposed method first on the task of taxonomy completion, 
before later using it to generate taxonomies without seed relations. 
In addition to gold-standard comparisons, we rely on recently introduced reference-free metrics 
which evaluate the robustness and logical adequacy of generated taxonomies \cite{wullschleger2025goldstandardproblemreferencefree}.






In summary, the contributions of this study are novel LLM-based algorithms for 1) taxonomy completion and 2) taxonomy generation given a set of potentially incomplete known concepts. In a comparison to state-of-the-art methods on five taxonomies, we demonstrate the potential of these algorithms for food-related and other taxonomies. Our implementations and datasets are publicly available on GitHub to ensure reproducibility\footnote{\url{https://github.com/wullli/foodtaxo}}.

\section{Related Work}

The task of taxonomy expansion was introduced as adding leaves to an existing taxonomy \cite{shen_2018, fauceglia-etal-2019-automatic, shen_2020, yu_2020, manzoor-2020, ma-etal-2021-hyperexpan-taxonomy, margiotta_2023}. However, \citet{Zhang_Song_Zeng_Chen_Shen_Mao_Li_2021} later argued that this is problematic, since it assumes that all newly extracted concepts are hyponyms of existing leaves in the taxonomy. To overcome this assumption, they present a triplet-matching approach, where they predict placements of query concepts as triplets of the form (parent, query, child). This new approach, termed taxonomy completion, allows for new concepts to be included as either hyponyms or hypernyms of existing concepts.

\citet{zeng_2021} formulate an extension to the taxonomy completion task whereby hypernym-hyponym pairs are not explicitly estimated, but candidate positions that require the addition of a new concept are identified. They argue that new concepts should not be extracted, but rather generated, since they can be rare and hard to extract in large text corpora. They initially predict the position in the taxonomy where a concept is missing, and subsequently generate the name of the concept given its position.

In contrast to \citet{zeng_2021}, our method does not require a seed
taxonomy for training, making it
applicable to generating taxonomies
solely based on a set of known concepts.
We make use of LLMs to generate and place concepts, whereas they train a \gls{GRU}-based decoder on the seed taxonomy to generate the names of concepts.

\citet{xu-etal-2023-tacoprompt} show few-shot prompting for taxonomy completion to be subpar to their prompt learning method (TacoPrompt). However, aside from few-shot examples, and in contrast to our proposed approach, they do not provide the model with relevant parts of the taxonomy as context. We compare  to TacoPrompt in Section~\ref{sec:experiments}.

\citet{chen_llm_tax_construct_2023} construct a taxonomy by determining hypernym-hyponym relationships among a set of concepts provided to an \gls{LLM}, demonstrating that prompt-based methods surpass fine-tuning, particularly as the size of the training taxonomy decreases. However, given the different setting, i.e. constructing a taxonomy using a complete concept set, a direct comparison with our approach is challenging.
\newcommand{\T}{\mathcal{T}}
\newcommand{\E}{\mathcal{E}}
\newcommand{\V}{\mathcal{V}}
\newcommand{\C}{\mathcal{C}}
\newcommand{\mL}{\mathcal{L}}

\section{Methodology}

\subsection{Problem Definition}
Following \citet{zeng_2021}, a taxonomy $\T = (\E, \V)$ is a 
directed acyclic graph with edges $(c_p, c_s) \in \E$ pointing 
from a parent vertex $c_p \in \V$ to a child vertex $c_s \in \V$. In the context of taxonomies, vertices are referred to as \textit{concepts}.
Edges represent hypernym-hyponym relations, where the child concept is the 
least detailed but different specialization of the parent concept.

Unlike traditional approaches \cite{shen_2020,manzoor-2020,Zhang_Song_Zeng_Chen_Shen_Mao_Li_2021,xu-etal-2023-tacoprompt} that assume a complete set of new concepts $\mathcal{Q}$ to be added to $\T$ to obtain a new taxonomy $\T' = (\E', \V \cup \mathcal{Q})$, we assume $\mathcal{Q}$ to be incomplete and allow for the generation of new concepts. 
Instead of starting with a fixed concept extraction process, we initialize $\mathcal{Q}$ with an incomplete set of known concepts (often leaves) that we want to categorize and iteratively insert  into the taxonomy with new 
concepts generated as needed. 

\citet{shen_2020}, \citet{manzoor-2020}, \citet{Zhang_Song_Zeng_Chen_Shen_Mao_Li_2021}, and \citet{xu-etal-2023-tacoprompt} assume for simplicity that adding a concept is independent of the attachment of other concepts, resulting in the irrelevance of the order of concept insertion. We observe that we can formulate the task of taxonomy generation as a recursive 
taxonomy completion task, where we remove the above independence assumption.
We start from an initial seed taxonomy $\T_0 = (\{\}, \V = \mathcal{Q} \cup \{p_l, p_r\})$ 
and iteratively predict placements for each $c \in \V$. 
A placement is a triplet $(c_p, c_q, c_s)$, where $c_q$ 
is the query concept that is placed as a child of $c_p$ and as a parent of $c_s$.
Following \citet{manzoor-2020}, we add a pseudo-leaf $p_l$ and pseudo-root $p_r$ to $T$ 
to allow insertion of concepts without parents or children. 
This means that if $c_q$ is inserted as a leaf, $c_s$ will be the pseudo-leaf node, and if $c_q$ is the root, then $c_p$ is the pseudo-root.
Note that $c_p$ can be either an existing concept in $\mathcal{Q}$ or a generated concept. 
If $c_p$ does not exist in $\mathcal{Q}$, we add it and predict its placement as well, 
thereby constructing the taxonomy in a bottom-up fashion using completions (Fig.~\ref{fig:intro}).

\subsection{Completing Taxonomies}
\label{sec:completing}

When completing a taxonomy, it, by definition, grows. Due to this, one 
cannot simply encode the whole tree into a string and use it as context 
in an \gls{LLM}, since a ceiling for sequence length would eventually be 
reached. Instead, we make use of well established techniques, such as chain-of-thought
 prompting \cite{wei_2022} and \gls{RAG} \cite{rag} as an initial retrieval step to provide 
the model with only the most relevant part of the taxonomy in order to insert the 
current query concept. For this purpose, we rely on the \gls{DSP} paradigm
\cite{khattab2023demonstratesearchpredictcomposingretrievallanguage}. 

The algorithm can be summarized as follows:
%
for each concept, $q \in \mathcal{Q}$, to insert: 
\begin{enumerate}[itemsep=0.2em]
\item Retrieve the most similar edges (parent, child) to $q$ based on cosine similarity using 
FastText embeddings\footnote{\url{https://dl.fbaipublicfiles.com/fasttext/vectors-crawl/cc.en.300.bin.gz}} \cite{bojanowski-etal-2017-enriching}.\footnote{For more detail on how the edges and concepts are encoded as strings, refer to the prompts in Appendix \ref{sec:prompts}.} 
\item Using \gls{CoT} prompting, retrieve potential candidates for parent concepts of $q$. In the completion case, these concepts are required to be in the set of existing seed (training) concepts. In case they are not, we repeat the prompt with additional information that the proposed concepts are not valid predictions. We call this backtracking.
In the generation case, we allow the model to invent non-existent concepts.
\item Subsequently retrieve the existing children of the proposed parents and again apply \gls{CoT} prompting to decide which of these children should be attached to the inserted concept.
\item Return all predicted placements as triplets of the form (parent, query, child).
\end{enumerate}
For more detail, see Algorithm \ref{alg:completion} in the Appendix.

\subsection{Generating Taxonomies}
We generate a taxonomy without a seed by initializing $\mathcal{Q}$ with a set of known concepts. These are the concepts we want to be able to classify using the taxonomy. Imagine a dataset of cooking recipes. We might want to classify all ingredients into a taxonomy to enable us to easily substitute an ingredient with one of its siblings. However, the set of concepts is unlikely to be complete when it is only initialized with ingredients. Broader concepts, such as \textit{dairy} will presumably not appear as an ingredient. Due to this, our model should predict possibly non-existent parents and children for known concepts, which will be added to the set and subsequently sent to the model for insertion into the current taxonomy. Thereby we construct a taxonomy in a bottom-up procedure. The following is a summary of the steps involved in the algorithm. For a more detailed description refer to Algorithm \ref{alg:generation} in the Appendix.

\begin{enumerate}
\item Initialize $\mathcal{Q}$, the set of concepts to insert, with all currently known concepts.
\item Sample 100 nodes from $\mathcal{Q}$ and let the LLM write a paragraph on what a potential taxonomy could look like (see, for example, the  \textit{Taxonomy Description} in App. \ref{sec:gen_prompt_app_p}).
\item While $\mathcal{Q}$ is not empty, do the following.
\begin{enumerate}[leftmargin=1em]
\item Perform the steps described in Section~\ref{sec:completing} for the completion case to insert $q \in \mathcal{Q}$ into the current taxonomy. 
\item If a new concept is generated, add it to $\mathcal{Q}$.
\item Update the taxonomy by inserting all predicted placements into the taxonomy graph.
\item Remove $q$ from the set of concepts, 
$Q$.
\end{enumerate}
\end{enumerate}

We may not end up with a single root node in the taxonomy. In such cases, all concepts without parents are attached to a pseudo-root. Ideally, the model predicts the pseudo-root as the parent of root nodes, providing a natural stopping criterion.

\subsection{Prompts} 
\label{sec:prompts}
We show handcrafted prompts for generating parent (Listing~\ref{lst:parent_prompt}) and child (Listing~\ref{lst:child_prompt}) concepts  of a query. We optimized these prompts by manual trial and error on the validation data. For spacing reasons, we display only two lines of context for both prompts and remove double new-lines.\ Note that the prompts for generation are slightly different.\ For a complete impression of prompts and model outputs, please consult Appendix \ref{sec:prompts_app}.

\begin{figure}
\begin{lstlisting}[linewidth=\linewidth,language=prompt]
// Prompt
Which are the most specific parent concepts of the given child concept in a taxonomy considering the context?
In your reasoning, state how the parent concepts are a supertype of the child concept.
Do not add additional comments or information, only return the output in the described format.

---

Input description.

Context: List of existing parent-child (supertype-subtype) relations in the taxonomy.

Child: Child concept (subtype) that you need to place in a taxonomy.

Description: Description of the child concept.

---

Follow the following format.

Reasoning: Let's think step by step in order to ${produce the parents}. We ...

Interpretation: Description of the child concept in relation to the context taxonomy. Infer what is meant by the child concept from the context.

Parents: Comma separated list of one or more parents of the child concept. Valid parents are in the context. If there are no suitable parents, return None.

---

Context:
```liqueur, sambuca
sugar, sugarloaf
sweet, hardbake
food, comestible
wine, riesling
liqueur, galliano
irish, poteen
flavorer, sassafras
dish, bitok
wine, sauterne
dish, kishke
starches, bap
condiment, chowchow
liqueur, pernod
wine, dubonnet
feed, eatage
cider, scrumpy
dish, rijsttaffel
wine, tokay
...
```

Child: sweetening

Description: sweetening is something added to foods to make them taste sweeter

Reasoning: Let's think step by step in order to


\end{lstlisting}%
\begin{lstlisting}[linewidth=\linewidth,language=prompt,style=generated,caption={Parent generation CoT prompt for the completion task. In the generation settings, this prompt looks slightly different (see Appendix \ref{sec:prompts_app})},captionpos=b,label={lst:parent_prompt}]
// Output
Reasoning: Let's think step by step in order to find the parents of the child concept "sweetening". We can infer that sweetening is related to food and its taste, so we need to find the concepts in the context that are related to food and taste.

Interpretation: The description of the child concept "sweetening" implies that it is an additive that enhances the sweetness of food, which is a type of comestible.

Parents: flavorer, condiment
\end{lstlisting}
\end{figure}

\begin{figure}
\begin{lstlisting}[linewidth=\linewidth,language=prompt]
// Prompt
Which of the candidates are child concepts (subtypes) of the given parent concept (supertype) in a taxonomy?
The context shows existing parent and child concepts and whether the children are leaves.
In your reasoning, state how the parent concept is a supertype of the selected child concepts.
Do not add additional comments or information, only return the output in the described format.

---

Input description.

Context: List of existing parent-child (supertype-subtype) relations in the taxonomy.

Candidates: Candidate children of the concept separated by commas to select from.

Parent: Parent concept that you need to place in a taxonomy.

Description: Description of the parent concept.

Interpretation: Description of the child concept in relation to the taxonomy.

---

Follow the following format.

Reasoning: Let's think step by step in order to ${produce the children}. We ...

Leaf: Whether the parent concept should be added as a leaf (has no children). Answer with Yes or No.

Children: Comma separated list of candidates that are children of the parent concept in a taxonomy.A child concept must be a type of the parent concept.Separate with commas.

---

Context:
```
...
```

Candidates: salsa, cranberry sauce, dip, soy sauce, wasabi, vinegar, spread, duck sauce, chutney, marinade, mustard, sauce, mint sauce, green olive, pickle relish, black olive, chowchow, pickle

Parent: sweetening

Description: sweetening is something added to foods to make them taste sweeter

Interpretation: The description of the child concept "sweetening" implies that it is an additive that enhances the sweetness of food, which is a type of comestible.

Reasoning: Let's think step by step in order to

\end{lstlisting}%
\begin{lstlisting}[linewidth=\linewidth,language=prompt,style=generated,caption={Child generation CoT prompt. Prompts are identical for the completion and generation cases.},captionpos=b,label={lst:child_prompt}]
// Output
Reasoning: Let's think step by step in order to identify the children of the concept "sweetening". We can see that sweetening is something added to foods to make them taste sweeter, which implies that it is a type of additive or condiment. Therefore, the child concepts should be types of sweetening agents.

Leaf: No

Children: sugar, salsa, cranberry sauce, dip, soy sauce, wasabi, vinegar, spread, duck sauce, chutney, marinade, mustard, sauce, mint sauce, pickle relish
\end{lstlisting}
\end{figure}

\subsection{Verifying Placements}
\label{sec:nli_verify}
We notice that sometimes an \gls{LLM} will mistakenly interpret the meaning of a parent-child relation as 
a similarity relation and simply generate predictions of related parent or child concepts on the basis 
of their similarity. This behaviour persisted even with careful instructions. 
To mitigate this, we require that the description of the concept (premise) entails the relation 
(hypothesis) when passed through an \gls{NLI} model for all predicted children. 
For the proposed parent concepts, the verification process is more lenient, requiring only the absence of contradictions. 
We examine the effect of this step in Section \ref{sec:gen_ablation}. 

\subsection{Backtracking}
\label{sec:backtrack_constraints}
To ensure valid model outputs, we re-prompt the models at most three times in case outputs do not fulfill constraints. For example, in the completion case, we require all predictions to be valid concepts that exist in the taxonomy, while this is disabled in generation mode so that we can generate suitable missing concepts. More specifically, we leverage the backtracking functionality provided with the \texttt{DSPy} library \cite{singhvi2024} in case any of the following assertions fail.

\begin{enumerate}[itemsep=0.2em]
\item The model predicts the query to be its own parent or child.
\item The model predicts non-existent parent concepts (completion case only).
\item The model predicts non-existent child concepts.
\item Parents are predicted, but none of them pass the NLI-verification. This does not apply if the model predicts the pseudo-root as a parent.
\item Children are predicted, but none of them pass the NLI-verification. This does not apply if the model predicts the pseudo-leaf as a child.
\item The concept consists of six or more words.
\item The model predicts children for a concept that are not actually present in the list of candidate children.
\end{enumerate}

\section{Completion Experiments}\label{sec:experiments}

\subsection{Data}
For benchmarking our completion approach, we follow \citet{xu-etal-2023-tacoprompt} and \citet{wang_2022} by evaluating on the SemEval-Food, SemEval-Verb and MeSH datasets. SemEval-Food is the largest taxonomy of the SemEval-2016 Task 13, that was used to evaluate taxonomy extraction methods for a given corpus \cite{bordea-etal-2016-semeval}. 
SemEval-Verb is based on WordNet 3.0 \cite{Fellbaum2010} and featured in the SemEval-2016 Task 14, which concerned evaluation of taxonomy enrichment approaches \cite{jurgens-pilehvar-2016-semeval}. 
MeSH is a hierarchically organized vocabulary of medical terms \cite{Lipscomb2000-wk}. 

Additionally, we extract a taxonomy from Wikidata\footnote{\url{https://www.wikidata.org/}} by selecting the data-item \texttt{Food (Q2095)} as the root node and extracting all children using the relations \textit{subclass of}, \textit{instances of} and \textit{subproperty of} (Wikidata identifiers P279, P31 and P1647). Lastly, we leverage a proprietary taxonomy provided by a large food market chain that is also being used for recipe development by Betty Bossi, a subsidiary publishing company specialized in consumer recipes. We call this the \textit{CookBook} taxonomy. Both taxonomies are available together with the source code.\footnote{\url{https://github.com/wullli/foodtaxo}}
\setlength\arrayrulewidth{1pt}
\begin{table}
    \centering
    \resizebox{0.48\textwidth}{!}
    {
    \setlength{\tabcolsep}{6pt}
    {
    \footnotesize\begin{tabular}{lrrrrrr}
        \toprule
         \textbf{Dataset} & $|\mathbfcal{V}|$ & $|\mathbfcal{E}|$& \textbf{$\mathbf{D}$} & \textbf{$\mathbf{|L|}$} & \textbf{$\mathbf{\frac{|L|}{|\mathbfcal{V}|}}$} & $\mathbf{B}$\\ 
         \midrule
        SemEval-Food & 1486 & 1576 & 9 & 1184 & 0.80 & 5.08 \\
        SemEval-Verb & 13936 & 13407 & 13 & 10360 & 0.74 & 4.12 \\
        MeSH & 9710 & 10496 & 11 & 5502 & 0.57 & 3.88 \\
        Wikitax & 941 & 973 & 7& 754 & 0.80 & 5.20 \\
        CookBook & 1985 & 1984 & 4 & 1795 & 0.90 & 10.44 \\
        \bottomrule
    \end{tabular}}
    }
  \caption{Statistics regarding the benchmark taxonomies. $|\mathcal{V}|$, $|\mathcal{E}|$, $D$, $|L|$, $\frac{|L|}{|\mathcal{V}|}$, $B$ represent the node number, edge number, depth, the number of leaves, the ratio of leaves and the branching factor of the taxonomy.}
  \label{tab:tax_stats} 
\end{table}

\vspace{1em}\begin{table}
    \centering
    \resizebox{0.48\textwidth}{!}
    {
\footnotesize\begin{tabular}{lrrr}
\toprule
\textbf{Dataset} & \textbf{Train} $|\mathbfcal{V}|$ & \textbf{Val} $|\mathbfcal{V}|$ & \textbf{Test} $|\mathbfcal{V}|$ \\
\midrule
SemEval-Food & 1190 (80.0\%) & 148 (10.0\%) & 148 (10.0\%) \\
SemEval-Verb & 11996 (86.0\%) & 1000 ( 7.0\%) & 1000 ( 7.0\%) \\
MeSH & 8072 (83.0\%) & 819 ( 8.0\%) & 819 ( 8.0\%) \\
Wikidata & 753 (80.0\%) & 94 (10.0\%) & 94 (10.0\%) \\
CookBook & 1589 (80.0\%) & 198 (10.0\%) & 198 (10.0\%) \\
\bottomrule
\end{tabular}}
\caption{Node counts per split and dataset for the completion evaluation.}
\label{tab:splits}
\end{table}

\setlength\arrayrulewidth{1pt}
\begin{table*}[!t]
    \centering
    \resizebox{1\textwidth}{!}
    {
    \footnotesize\begin{tabular}{llrrrrrrrrrrrrrr}
         \toprule
         \multirow{2}{*}{\textbf{Dataset}} & 
         \multirow{2}{*}{\textbf{Model}} &
         \multicolumn{4}{c}{\textbf{Total}} && 
         \multicolumn{4}{c}{\textbf{Non-Leaf}} && 
         \multicolumn{4}{c}{\textbf{Leaf}}\\
         \cmidrule{3-6}
         \cmidrule{8-11}
         \cmidrule{13-16}
         && WPS & F1 & P & R && WPS & F1 & P & R && WPS & F1 & P & R\\
         \midrule
         \multirow{10}{*}{SemEval-Food}
            & Arborist & 0.7184 & 0.0828 & 0.1284 & 0.0611 &  & 0.7794 & 0.0199 & 0.0800 & 0.0114 &  & 0.7060 & 0.1318 & 0.1382 & 0.1259 \\
            & QEN & \underline{0.8900} & 0.2919 & 0.4527 & 0.2154 &  & 0.9042 & \underline{0.0498} & \underline{0.2000} & \underline{0.0284} &  & \underline{0.8871} & 0.4806 & 0.5041 & 0.4593 \\
            & TEMP & \underline{0.8945} & \underline{0.3529} & \underline{0.5473} & \underline{0.2605} &  & \underline{0.9155} & \underline{0.0896} & \underline{0.3600} & \underline{0.0511} &  & \underline{0.8902} & \underline{0.5581} & \underline{0.5854} & \underline{0.5333} \\
            & TMN & 0.8226 & 0.1089 & 0.1689 & 0.0804 &  & 0.8365 & \underline{0.0299} & \underline{0.1200} & \underline{0.0170} &  & 0.8198 & 0.1705 & 0.1789 & 0.1630 \\
            & TacoPrompt & \textbf{0.9054} & \textbf{0.4052} & \textbf{0.6284} & \textbf{0.2990} &  & \textbf{0.9603} & \textbf{0.0995} & \underline{0.4000} & \textbf{0.0568} &  & \textbf{0.8942} & \textbf{0.6434} & \textbf{0.6748} & \textbf{0.6148} \\
            & TaxoExpan & 0.8021 & 0.0566 & 0.0878 & 0.0418 &  & 0.8288 & 0.0100 & 0.0400 & 0.0057 &  & 0.7967 & 0.0930 & 0.0976 & 0.0889 \\
            \cmidrule{2-16}
            & Llama-3 Few-Shot & 0.8560 & 0.3025 & 0.5076 & 0.2154 &  & 0.8168 & \underline{0.0914} & \textbf{0.4286} & \underline{0.0511} &  & \underline{0.8639} & 0.4715 & 0.5225 & 0.4296 \\
            & Llama-3 Zero-Shot & 0.8164 & 0.2192 & 0.3780 & 0.1543 &  & 0.8005 & \underline{0.0508} & \underline{0.2381} & \underline{0.0284} &  & 0.8196 & 0.3568 & 0.4057 & 0.3185 \\

        \midrule    
        \multirow{10}{*}{SemEval-Verb}
            & Arborist & 0.7430 & 0.0000 & 0.0000 & 0.0000 &  & 0.7359 & 0.0000 & 0.0000 & 0.0000 &  & 0.7437 & 0.0000 & 0.0000 & 0.0000 \\
            & QEN & \textbf{0.8321} & 0.0967 & 0.1205 & 0.0808 &  & \textbf{0.8624} & \underline{0.0056} & \underline{0.0127} & \underline{0.0036} &  & \textbf{0.8292} & 0.1167 & 0.1323 & 0.1044 \\
            & TEMP & \underline{0.8184} & 0.1431 & 0.1782 & 0.1195 &  & 0.8146 & \underline{0.0224} & \underline{0.0506} & \underline{0.0144} &  & \underline{0.8187} & \underline{0.1695} & \underline{0.1922} & \underline{0.1516} \\
            & TMN & 0.8036 & 0.0081 & 0.0100 & 0.0067 &  & \underline{0.8276} & \underline{0.0056} & \underline{0.0127} & \underline{0.0036} &  & 0.8012 & 0.0086 & 0.0097 & 0.0077 \\
            & TacoPrompt & \underline{0.8242} & \textbf{0.1652} & \textbf{0.2058} & \textbf{0.1380} &  & \underline{0.8607} & \textbf{0.0392} & \textbf{0.0886} & \textbf{0.0252} &  & \underline{0.8207} & \textbf{0.1929} & \textbf{0.2187} & \textbf{0.1725} \\
            & TaxoExpan & 0.7896 & 0.0161 & 0.0201 & 0.0135 &  & 0.7756 & 0.0000 & 0.0000 & 0.0000 &  & 0.7910 & 0.0197 & 0.0223 & 0.0176 \\
            \cmidrule{2-16}
            & Llama-3 Few-Shot & 0.7879 & 0.0630 & 0.0814 & 0.0513 &  & \underline{0.8332} & \underline{0.0113} & \underline{0.0263} & \underline{0.0072} &  & 0.7835 & 0.0745 & 0.0877 & 0.0648 \\
            & Llama-3 Zero-Shot & 0.7792 & 0.0608 & 0.0784 & 0.0497 &  & 0.8019 & \underline{0.0113} & \underline{0.0267} & \underline{0.0072} &  & 0.7770 & 0.0718 & 0.0841 & 0.0626 \\
        
        \midrule
        \multirow{10}{*}{MeSH}
        & Arborist & 0.5131 & 0.0000 & 0.0000 & 0.0000 &  & 0.5394 & 0.0000 & 0.0000 & 0.0000 &  & 0.5008 & 0.0000 & 0.0000 & 0.0000 \\
        & QEN & \underline{0.8609} & 0.1181 & 0.1978 & 0.0842 &  & 0.8815 & 0.0385 & 0.1077 & 0.0234 &  & \textbf{0.8513} & 0.2081 & 0.2397 & 0.1838 \\
        & TEMP & 0.8311 & 0.1866 & 0.3126 & 0.1330 &  & 0.8686 & 0.0742 & 0.2077 & 0.0452 &  & 0.8137 & 0.3137 & 0.3614 & 0.2771 \\
        & TMN & 0.5241 & 0.0000 & 0.0000 & 0.0000 &  & 0.5515 & 0.0000 & 0.0000 & 0.0000 &  & 0.5114 & 0.0000 & 0.0000 & 0.0000 \\
        & TacoPrompt & \textbf{0.8613} & \textbf{0.2201} & \underline{0.3687} & \textbf{0.1569} &  & \textbf{0.9070} & 0.0673 & 0.1885 & 0.0410 &  & \underline{0.8401} & \textbf{0.3929} & \textbf{0.4526} & \textbf{0.3471} \\
        & TaxoExpan & 0.5194 & 0.0020 & 0.0202 & 0.0010 &  & 0.5494 & 0.0000 & 0.0000 & 0.0000 &  & 0.5054 & 0.0051 & 0.0351 & 0.0027 \\
        \cmidrule{2-16}
        & Llama-3 Few-Shot & \underline{0.8509} & \underline{0.2139} & \textbf{0.3750} & \underline{0.1496} &  & 0.8616 & \textbf{0.1126} & \textbf{0.3333} & \textbf{0.0677} &  & \underline{0.8459} & 0.3301 & \underline{0.3943} & 0.2840 \\
        & Llama-3 Zero-Shot & \underline{0.8481} & 0.1662 & 0.2877 & 0.1169 &  & 0.8563 & 0.0845 & 0.2460 & 0.0510 &  & \underline{0.8444} & 0.2597 & 0.3071 & 0.2250 \\
            
        \midrule
        \multirow{10}{*}{Wikidata}
        & Arborist & 0.7865 & 0.0556 & 0.0638 & 0.0492 &  & 0.7467 & 0.0000 & \underline{0.0000} & \underline{0.0000} &  & 0.7935 & 0.0741 & 0.0750 & 0.0732 \\
        & QEN & \underline{0.8663} & 0.1574 & 0.1809 & 0.1393 &  & \underline{0.8143} & \underline{0.0370} & \underline{0.0714} & \underline{0.0250} &  & \underline{0.8754} & \underline{0.1975} & \underline{0.2000} & \underline{0.1951} \\
        & TEMP & 0.8513 & \underline{0.2593} & \underline{0.2979} & \underline{0.2295} &  & \underline{0.8710} & \underline{0.1111} & \underline{0.2143} & \underline{0.0750} &  & 0.8479 & \underline{0.3086} & \underline{0.3125} & \underline{0.3049} \\
        & TMN & 0.7973 & 0.0926 & 0.1064 & 0.0820 &  & 0.7650 & \underline{0.0370} & \underline{0.0714} & \underline{0.0250} &  & 0.8029 & 0.1111 & 0.1125 & 0.1098 \\
        & TacoPrompt & \textbf{0.8888} & \underline{0.2130} & \underline{0.2447} & \underline{0.1885} &  & \textbf{0.8882} & \underline{0.1111} & \underline{0.2143} & \underline{0.0750} &  & \underline{0.8889} & \underline{0.2469} & \underline{0.2500} & \underline{0.2439} \\
        & TaxoExpan & 0.7818 & 0.0185 & 0.0213 & 0.0164 &  & \underline{0.8599} & 0.0000 & \underline{0.0000} & \underline{0.0000} &  & 0.7682 & 0.0247 & 0.0250 & 0.0244 \\
        \cmidrule{2-16}
        & Llama-3 Few-Shot & \underline{0.8864} & \textbf{0.2870} & \textbf{0.3298} & \textbf{0.2541} &  & \underline{0.8465} & \textbf{0.1481} & \textbf{0.2857} & \textbf{0.1000} &  & \textbf{0.8934} & \textbf{0.3333} & \textbf{0.3375} & \textbf{0.3293} \\
        & Llama-3 Zero-Shot & \underline{0.8744} & \underline{0.2407} & \underline{0.2766} & \underline{0.2131} &  & \underline{0.8166} & \underline{0.1111} & \underline{0.2143} & \underline{0.0750} &  & \underline{0.8845} & \underline{0.2840} & \underline{0.2875} & \underline{0.2805} \\

        \midrule
        \multirow{10}{*}{CookBook}
        & Arborist & 0.8536 & 0.0156 & 0.0202 & 0.0127 &  & 0.8743 & \underline{0.0253} & \underline{0.1000} & \underline{0.0145} &  & 0.8513 & 0.0112 & 0.0112 & 0.0112 \\
        & QEN & \underline{0.9099} & 0.1868 & 0.2424 & 0.1519 &  & \underline{0.9086} & \underline{0.0253} & \underline{0.1000} & \underline{0.0145} &  & \underline{0.9101} & 0.2584 & 0.2584 & 0.2584 \\
        & TEMP & \underline{0.9206} & 0.2529 & 0.3283 & \underline{0.2057} &  & \underline{0.9452} & \underline{0.0506} & \underline{0.2000} & \underline{0.0290} &  & \underline{0.9179} & 0.3427 & 0.3427 & 0.3427 \\
        & TMN & 0.8495 & 0.0623 & 0.0808 & 0.0506 &  & \underline{0.8990} & \underline{0.0253} & \underline{0.1000} & \underline{0.0145} &  & 0.8439 & 0.0787 & 0.0787 & 0.0787 \\
        & TacoPrompt & \underline{0.9243} & \underline{0.2879} & \underline{0.3737} & \underline{0.2342} &  & \underline{0.9300} & \underline{0.0506} & \underline{0.2000} & \underline{0.0290} &  & \underline{0.9236} & \underline{0.3933} & \underline{0.3933} & \underline{0.3933} \\
        & TaxoExpan & 0.8234 & 0.0272 & 0.0354 & 0.0222 &  & 0.7713 & \underline{0.0127} & \underline{0.0500} & \underline{0.0072} &  & 0.8293 & 0.0337 & 0.0337 & 0.0337 \\
        \cmidrule{2-16}
        & Llama-3 Few-Shot & \textbf{0.9342} & \textbf{0.3327} & \textbf{0.4359} & \textbf{0.2690} &  & \textbf{0.9629} & \textbf{0.0633} & \textbf{0.2500} & \textbf{0.0362} &  & \textbf{0.9310} & \textbf{0.4533} & \textbf{0.4571} & \textbf{0.4494} \\
        & Llama-3 Zero-Shot & 0.9089 & 0.2383 & 0.3112 & 0.1930 &  & \underline{0.9343} & \underline{0.0380} & \underline{0.1500} & \underline{0.0217} &  & 0.9060 & 0.3277 & 0.3295 & 0.3258 \\

        \bottomrule
    \end{tabular}
    }
  \caption{Scores of the completion evaluation on all datasets. All scores that are not significantly different to the best model according to a two-sided paired randomization test ($\alpha = 0.05$) with 1k resamples are underlined. Note that due to the rarity of non-leaves, these results rarely show significant differences.}
  \label{tab:eval} 
\end{table*}

\begin{table*}[h!]
    \centering
    \resizebox{1\textwidth}{!}
    {
    \footnotesize\begin{tabular}{llrrrrrrrrrrrrrr}
         \toprule
         \multirow{2}{*}{\textbf{Setting}} & 
         \multirow{2}{*}{\textbf{Model}} &
         \multicolumn{4}{c}{\textbf{Total}} && 
         \multicolumn{4}{c}{\textbf{Non-Leaf}} && 
         \multicolumn{4}{c}{\textbf{Leaf}}\\
         \cmidrule{3-6}
         \cmidrule{8-11}
         \cmidrule{13-16}
         && WPS & F1 & P & R && WPS & F1 & P & R && WPS & F1 & P & R\\
         \midrule
         \multirow{3}{*}{Zero-Shot}
            & w/o Backtracking & \underline{0.7970} & \textbf{0.2454} & \textbf{0.4380} & \textbf{0.1704} &  & \underline{0.7654} & \underline{0.0306} & \underline{0.1500} & \underline{0.0170} &  & \underline{0.8034} & \textbf{0.4237} & \textbf{0.4950} & \textbf{0.3704} \\
            & w/o NLI Validation & \textbf{0.8206} & \underline{0.2257} & \underline{0.3788} & \underline{0.1608} &  & \underline{0.7567} & \underline{0.0406} & \underline{0.1905} & \underline{0.0227} &  & \textbf{0.8336} & \underline{0.3740} & \underline{0.4144} & \underline{0.3407} \\
            \cmidrule{2-16}
            & Complete & \underline{0.8175} & \underline{0.2192} & \underline{0.3780} & \underline{0.1543} &  & \textbf{0.8027} & \textbf{0.0508} & \textbf{0.2381} & \textbf{0.0284} &  & \underline{0.8205} & \underline{0.3568} & \underline{0.4057} & \underline{0.3185} \\
        \midrule
         \multirow{3}{*}{Few-Shot}
            & w/o Backtracking & 0.8052 & \underline{0.2593} & \underline{0.4628} & \underline{0.1801} &  & 0.7616 & \underline{0.0622} & \underline{0.3529} & \underline{0.0341} &  & 0.8140 & \underline{0.4184} & \underline{0.4808} & \underline{0.3704} \\
            & w/o NLI Validation & \underline{0.8581} & \underline{0.2793} & \underline{0.4662} & \underline{0.1994} &  & \underline{0.8175} & \underline{0.0711} & \underline{0.3333} & \underline{0.0398} &  & \textbf{0.8664} & \underline{0.4453} & \underline{0.4911} & \underline{0.4074} \\
            \cmidrule{2-16}
            & Complete & \textbf{0.8583} & \textbf{0.3025} & \textbf{0.5076} & \textbf{0.2154} &  & \textbf{0.8282} & \textbf{0.0914} & \textbf{0.4286} & \textbf{0.0511} &  & \underline{0.8645} & \textbf{0.4715} & \textbf{0.5225} & \textbf{0.4296} \\
        \bottomrule
    \end{tabular}
    }
  \caption{Ablation study of NLI-verification and Backtracking on the completion task for SemEval-Food. All scores that are not significantly different to the best model according to a two-sided paired randomization test ($\alpha = 0.05$) with 1k resamples are underlined.}
  \label{tab:ablation_completion} 
\end{table*}

\subsection{Evaluation}
Due to our generative approach, we do not return a ranked list of candidate positions, making ranking metrics inappropriate for 
our case. Thus only precision (P),  recall (R) and F1-scores (F1) of candidate positions (parent-query-child triplets) that were generated during inference are calculated.  Following \citet{liu-etal-2021-temp}, we additionally calculate the \gls{WPS} \cite{wu-palmer-1994-verb}. It measures the similarity between the paths in a taxonomy and is commonly known for its application as a similarity score with WordNet \cite{Fellbaum2010}. Let $p(c_t) = \langle c_r, ..., c_t\rangle$ be the path from the pseudo-root concept $c_r$ to a target concept $c_t$.\
Likewise, let $\mathtt{lca}(c_a, c_b)$ denote the depth of the least common ancestor of the nodes $c_a$ and $c_b$. The \gls{WPS} (Eq. \ref{eq:wps}) represents the similarity between concepts $c_a$ and $c_b$ where $p(c_a)$ and $p(c_b)$ are the paths from the root node to $c_a$ and $c_b$. The score ranges $(0, 1]$, with $1$ meaning that they share a parent.

\begin{eqnarray}
WPS_{c_a c_b} &=& \frac{2 \cdot \mathtt{lca}(c_a, c_b)}{\lvert p(c_a)\rvert + \lvert p(c_b)\rvert}
\label{eq:wps}
\end{eqnarray}

We follow \citet{wang_2022} in splitting the benchmark datasets into train (seed), validation and test taxonomies. 
We randomly exclude nodes (except root) and connect parents of excluded nodes with their children to keep the training (seed) taxonomy intact.
An overview of the node counts per split can be found in Table~\ref{tab:splits}. 

In order to gain insights into performances across different node types, we provide total scores, as well as leaf and non-leaf scores. 
The leaf scores are a proxy for the performances on a taxonomy expansion task, where only leaves must be added.

\paragraph{Model selection} 
Since running experiments on LLMs is expensive, and we want to make our approach easily accessible, we restrict our experiments to the open-source model Llama-3 (\texttt{Llama-3-70b-Instruct}).\footnote{\url{https://huggingface.co/meta-llama/Meta-Llama-3-70B-Instruct}}


\paragraph{Hypothesis testing}
Following the recommendations of \citet{dror-etal-2018-hitchhikers}, we use a two-sided paired randomization test ($\alpha = 0.05$) with 1k resamples to assess significant differences in model performance in the completion experiments. Since listing all $p$-values would require tables with hundreds of rows, we refrain from adding them here. However, they can be calculated using our published source code.

\subsection{Results}

Table~\ref{tab:eval} shows that LLM-based taxonomy completion can be competitive with state-of-the-art methods, 
even without tuning. The LLM approach is competitive with previous approaches on 3 of the 5 evaluated datasets. It is the best performing method on the CookBook taxonomy. However, it performs rather poorly on SemEval-Verb, 
the largest of the benchmark taxonomies -- it is possible that fine-tuning becomes more advantageous as the size of the taxonomy increases. In all cases, few-shot prompting outperforms zero-shot, although the differences are not always statistically significant.

We further experimented with methods to automatically tune the prompt texts, but observed no significant difference to our manually optimized prompt. For details consult Appendix~\ref{sec:dspy}.

\paragraph{Ablations}
In order to justify the usage of backtracking and NLI-verification, we evaluated ablated versions of the method on SemEval-Food (Table~\ref{tab:ablation_completion}). Improvements are inconsistent overall, except for the non-leaf case, where the unablated model performs best for both zero-shot and few-shot. However, the scores are not significantly different according to randomization tests. 
\section{Generation Experiments}

\subsection{Data}
To facilitate direct comparisons between true, generated and completed taxonomies, we extract all leaf-concepts from MeSH and SemEval-Food and try to regenerate a taxonomy only based on these known concepts. 


\subsection{Evaluation}
Instead of only comparing our generated taxonomy to a 
gold standard, we acknowledge that there may be multiple valid taxonomies based on an single initial set of known concepts. Therefore, we additionally assess the taxonomies 
using reference-free metrics \cite{wullschleger2025goldstandardproblemreferencefree}. In particular we evaluate \gls{CSC} and \gls{NLIV}, 
and compare scores between the generated and benchmark taxonomies. 

\gls{CSC} measures taxonomy robustness by correlating the taxonomic similarity of concepts (using WPS) with their semantic similarities according to an embedding model. Robustness indicates how well a taxonomy can tell things apart, meaning how clearly the concepts in a taxonomy represent different ideas (orthogonality) and how closely related sibling concepts are (cohesiveness).

\gls{NLIV} evaluates logical adequacy by checking the validity of relations in a taxonomy. More specifically, if the process of classification is a walk on a taxonomy graph (from root node to classified node), then \gls{NLIV} estimates classification probabilities with \gls{NLI} and normalizes them by walk length. For example, in a food taxonomy, given the relation (\textit{antipasto}, \textit{appetizer}), the premise \textit{"antipasto is a course of appetizers in an Italian meal"} and hypothesis \textit{"antipasto is a kind of appetizer"} are passed to an \gls{NLI}-model. NLIV has two versions: weak (NLIV-W), where the premise must not contradict the hypothesis, and strong (NLIV-S), where the premise must entail it. Note that due to our model-internal NLI-verification (see Section \ref{sec:nli_verify}), results might be biased towards our model. However, we use two unrelated \gls{NLI}-models for evaluation and completion to improve fairness (see Appendix~\ref{sec:nli_verify_app}).

\paragraph{Gold-Standard Comparison}
For reference, we also calculate F1-scores on the complete gold standard taxonomy, which indicate how much of the gold standard was recovered during generation. The Position-F1 indicates how many triplets were matched, while the Parent-F1 indicates how often the correct parent, but not child, was predicted.

\subsection{Results}

\setlength\arrayrulewidth{1pt}
\begin{table}[t]
    \centering
    \resizebox{0.48\textwidth}{!}
    {
\footnotesize\begin{tabular}{llrrrrrr}
\toprule
\textbf{Dataset} & \textbf{Taxonomy} & 
\multicolumn{2}{c}{\textbf{vs. Gold Standard}}
 && \multicolumn{3}{c}{\textbf{Reference-free}} \\
\cmidrule{3-4}
\cmidrule{6-8}
&& Position-F1 & Parent-F1 && NLIV-W & NLIV-S & CSC\\
\midrule
\multirow{5}{*}{SemEval-Food}
&TacoPrompt & \underline{0.6432} & \textbf{0.7249} && 0.3479 & 0.0451 & -0.0023 \\
&True & - & - && 0.9641 & \textbf{0.2017} & \underline{0.0426} \\
\cmidrule{2-8}
&Completed & \textbf{0.6435} & \underline{0.7159} && 0.9525 & \underline{0.1774} & 0.0097 \\
&Generated & 0.0234 & 0.0390 && \textbf{0.9726} & 0.1298 & \textbf{0.0777} \\
\midrule
\multirow{5}{*}{MeSH}
&TacoPrompt & \textbf{0.6584} & \textbf{0.7397} && 0.5638 & 0.0510 & 0.0050 \\
&True & - & - && \textbf{0.8502} & \textbf{0.1680} & \underline{0.0614} \\
\cmidrule{2-8}
&Completed & \underline{0.6368} & \underline{0.7275} && \underline{0.8412} & \underline{0.1560} & 0.0518 \\
&Generated & 0.0094 & 0.0175 && 0.8167 & 0.1237 & \textbf{0.1051} \\
\bottomrule
\end{tabular}}
  \caption{Comparison of metrics for the true taxonomy, completed taxonomy (Ours and TacoPrompt) and a taxonomy constructed by our generation method.}
  \label{tab:gen_comparison} 
\end{table}

\begin{figure}[H]
\centering
\subfloat[]{\label{fig:sefood_expl}
\includegraphics[width=0.55\linewidth] {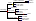}}
\subfloat[]{\label{fig:gen_expl1}
\includegraphics[width=0.44\linewidth] {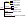}}\vspace{-1em}\\
\subfloat[]{\label{fig:gen_expl2}
\includegraphics[width=0.7\linewidth] {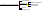}}
\label{fig:example_generated}
\caption{Examples of generated sub-graphs of the taxonomies. Depiction (a) is the gold standard neighborhood of \textit{coffee}, 
while (b) is an example of the generated taxonomy based on SemEval-Food leaves. An erroneously classified non-leaf is shown in (c).}
\end{figure}

Table~\ref{tab:gen_comparison} shows a comparison of our generation method against the gold standard, TacoPrompt and our completion method on SemEval-Food and MeSH. We can see that our 
reference-free
scores are competitive with the gold standard and according to \gls{CSC} even better on both datasets. However, the CSC score does not respect that there might be invalid relationships in the taxonomy (not of type \texttt{is-a}) and we find by qualitative inspection that NLIV better represents the actual quality of the taxonomy. Further, we notice that there are frequent erroneous classifications (example Fig.~\ref{fig:gen_expl2}), which 
are not 
well 
captured by the metrics. 
Such issues likely stem from poor model performance on non-leaves (Table~\ref{tab:eval}). Table~\ref{tab:tax_stats_gen} shows statistics regarding the generated taxonomies.

\setlength\arrayrulewidth{1pt}
\begin{table}[b]
    \centering
    \small
    \resizebox{0.48\textwidth}{!}
    {
    \setlength{\tabcolsep}{6pt}
    {
    \begin{tabular}{lrrrrrr}
        \toprule
         \textbf{Dataset} & $|\mathbfcal{V}|$ & $|\mathbfcal{E}|$& \textbf{$\mathbf{D}$} & \textbf{$\mathbf{|L|}$} & \textbf{$\mathbf{\frac{|L|}{|\mathbfcal{V}|}}$} & $\mathbf{B}$\\ 
         \midrule     
        MeSH & 6908 & 6858 & 10 & 5712 & 0.83 & 5.65 \\
        SemEval-Food & 1213 & 1257 & 11 & 1130 & 0.93 & 15.14 \\
        \midrule   
        SemEval-Food (w/o NLI) & 1203 & 1216 & 6 & 1122 & 0.93 & 15.01 \\
        SemEval-Food (w/o Backtracking) & 1228 & 1272 & 7 & 1108 & 0.90 & 10.60 \\
        SemEval-Food (w/o Generation) & 1233 & 1251 & 12 & 1135 & 0.92 & 12.77 \\
        \bottomrule
    \end{tabular}}
    }
  \caption{Statistics regarding generated taxonomies. $|\mathcal{V}|$, $|\mathcal{E}|$, $D$, $|L|$, $\frac{|L|}{|\mathcal{V}|}$, $B$ represent the node number, edge number, depth, the number of leaves, the ratio of leaves and the branching factor of the taxonomy.}
  \label{tab:tax_stats_gen} 
\end{table}

\paragraph{Ablations}
\label{sec:gen_ablation}
In order to test the effectiveness of our modeling choices, we conducted an ablation study by removing different mechanisms from our algorithm. In Table~\ref{tab:abl_gen} we present the results for models without NLI-verification, taxonomy description, backtracking, and generation. Without generation, only existing concepts can be used to build the taxonomy. In the configuration without a taxonomy description, we remove the initial step, where we let an LLM imagine a potential taxonomy. 

All of our mechanisms result in an improvement of either CSC or NLIV. We observe the best CSC score for the model without NLI-verification, but when qualitatively exploring the taxonomy generated by this model, we observe frequent cases where an edge does not represent an \textit{is-a} relation, which is better reflected in the NLIV score. 

\setlength\arrayrulewidth{1pt}
\begin{table}[t]
    \centering
    \resizebox{0.42\textwidth}{!}
    {
\footnotesize\begin{tabular}{lrrr}
\toprule
\textbf{Configuration}& \textbf{CSC} & \textbf{NLIV-S} & \textbf{NLIV-W}\\
\midrule
w/o NLI-Verification & \textbf{0.0785} & 0.1126 & 0.9630 \\
w/o Taxonomy Description & 0.0386 & 0.1140 & 0.9607 \\
w/o Generation & 0.0445 & \textbf{0.1519} & \underline{0.9717} \\
w/o Backtracking & 0.0328 & 0.1091 & 0.9683 \\
\midrule
Complete & \underline{0.0703} & \underline{0.1298} & \textbf{0.9726} \\
\bottomrule
\end{tabular}}
  \caption{Ablation study highlighting the effects of NLI validation and taxonomy description on the generation metrics. The study was done by constructing a taxonomy using all leaf concepts from SemEval-Food.}
  \label{tab:abl_gen} 
\end{table}

\section{Conclusion}
We introduce an algorithm for the generation of taxonomies given a set of known concepts using \glspl{LLM}, thereby enabling us to scale taxonomies to dataset sizes beyond what can be managed by human curators with sensible efforts. We benchmark our LLM-based approach against state-of-the-art taxonomy completion methods, demonstrating its potential. Despite the fact that our research endeavor stems from the food technology industry, the presented methods for taxonomy generation and completion are general and agnostic to the concrete use-case or industry. Some of our experiments therefore involve linguistic and healthcare taxonomies. 

The taxonomies generated by our method achieve promising scores across existing quality metrics. However, qualitative inspection reveals that they still fall short of the nuance seen in human-curated taxonomies. We conclude that for \gls{LLM}-based taxonomy generation to reach practical utility, significant advances are still needed, particularly in the reliable placement of non-leaf concepts.

\clearpage\section{Limitations}

\begin{itemize}
\item Due to the computational overhead associated with LLMs, our experiments are only carried out using one open-source LLM. Care should be taken when interpreting results based on one LLM alone.
\item Our current approach does not generate taxonomies with respect to a target application, which is important in practical scenarios. 
\item While reference-free metrics hint at taxonomy quality, they are likely non-exhaustive and always need to be assessed in combination, since they measure different properties of taxonomy quality.
\end{itemize}

\section{Acknowledgements}
We would like to express our sincere appreciation to Betty Bossi\footnote{\url{https://www.bettybossi.ch/}} for their support of this research project and for providing us with their taxonomy used for recipe development. This research is supported through computing resources by the ADAPT
Centre for Digital Content Technology, which is funded under the SFI Research Centres Programme (Grant 13/RC/2106\_P2) and is co-funded under the European Regional Development. 
The authors thank the reviewers for their insightful and helpful comments.
\bibliography{refs}

\begin{thebibliography}{29}
\providecommand{\natexlab}[1]{#1}

\bibitem[{Bojanowski et~al.(2017)Bojanowski, Grave, Joulin, and Mikolov}]{bojanowski-etal-2017-enriching}
Piotr Bojanowski, Edouard Grave, Armand Joulin, and Tomas Mikolov. 2017.
\newblock \href {https://doi.org/10.1162/tacl_a_00051} {Enriching word vectors with subword information}.
\newblock \emph{Transactions of the Association for Computational Linguistics}, 5:135--146.

\bibitem[{Bordea et~al.(2016)Bordea, Lefever, and Buitelaar}]{bordea-etal-2016-semeval}
Georgeta Bordea, Els Lefever, and Paul Buitelaar. 2016.
\newblock \href {https://doi.org/10.18653/v1/S16-1168} {{S}em{E}val-2016 task 13: Taxonomy extraction evaluation ({TE}x{E}val-2)}.
\newblock In \emph{Proceedings of the 10th International Workshop on Semantic Evaluation ({S}em{E}val-2016)}, pages 1081--1091, San Diego, California. Association for Computational Linguistics.

\bibitem[{Chen et~al.(2023)Chen, Yi, and Varró}]{chen_llm_tax_construct_2023}
Boqi Chen, Fandi Yi, and Dániel Varró. 2023.
\newblock \href {https://doi.org/10.1109/MODELS-C59198.2023.00097} {Prompting or fine-tuning? a comparative study of large language models for taxonomy construction}.
\newblock In \emph{2023 ACM/IEEE International Conference on Model Driven Engineering Languages and Systems Companion (MODELS-C)}, pages 588--596.

\bibitem[{Dror et~al.(2018)Dror, Baumer, Shlomov, and Reichart}]{dror-etal-2018-hitchhikers}
Rotem Dror, Gili Baumer, Segev Shlomov, and Roi Reichart. 2018.
\newblock \href {https://doi.org/10.18653/v1/P18-1128} {The hitchhiker`s guide to testing statistical significance in natural language processing}.
\newblock In \emph{Proceedings of the 56th Annual Meeting of the Association for Computational Linguistics (Volume 1: Long Papers)}, pages 1383--1392, Melbourne, Australia. Association for Computational Linguistics.

\bibitem[{Fauceglia et~al.(2019)Fauceglia, Gliozzo, Dash, Chowdhury, and Mihindukulasooriya}]{fauceglia-etal-2019-automatic}
Nicolas~Rodolfo Fauceglia, Alfio Gliozzo, Sarthak Dash, Md. Faisal~Mahbub Chowdhury, and Nandana Mihindukulasooriya. 2019.
\newblock \href {https://doi.org/10.18653/v1/D19-3005} {Automatic taxonomy induction and expansion}.
\newblock In \emph{Proceedings of the 2019 Conference on Empirical Methods in Natural Language Processing and the 9th International Joint Conference on Natural Language Processing (EMNLP-IJCNLP): System Demonstrations}, pages 25--30, Hong Kong, China. Association for Computational Linguistics.

\bibitem[{Fellbaum(2010)}]{Fellbaum2010}
Christiane Fellbaum. 2010.
\newblock \href {https://doi.org/10.1007/978-90-481-8847-5_10} {\emph{WordNet}}, pages 231--243.
\newblock Springer Netherlands, Dordrecht.

\bibitem[{Jurgens and Pilehvar(2016)}]{jurgens-pilehvar-2016-semeval}
David Jurgens and Mohammad~Taher Pilehvar. 2016.
\newblock \href {https://doi.org/10.18653/v1/S16-1169} {{S}em{E}val-2016 task 14: Semantic taxonomy enrichment}.
\newblock In \emph{Proceedings of the 10th International Workshop on Semantic Evaluation ({S}em{E}val-2016)}, pages 1092--1102, San Diego, California. Association for Computational Linguistics.

\bibitem[{Khattab et~al.(2023{\natexlab{a}})Khattab, Santhanam, Li, Hall, Liang, Potts, and Zaharia}]{khattab2023demonstratesearchpredictcomposingretrievallanguage}
Omar Khattab, Keshav Santhanam, Xiang~Lisa Li, David Hall, Percy Liang, Christopher Potts, and Matei Zaharia. 2023{\natexlab{a}}.
\newblock \href {https://arxiv.org/abs/2212.14024} {Demonstrate-search-predict: Composing retrieval and language models for knowledge-intensive nlp}.
\newblock \emph{Preprint}, arXiv:2212.14024.

\bibitem[{Khattab et~al.(2023{\natexlab{b}})Khattab, Singhvi, Maheshwari, Zhang, Santhanam, Vardhamanan, Haq, Sharma, Joshi, Moazam, Miller, Zaharia, and Potts}]{khattab2023dspycompilingdeclarativelanguage}
Omar Khattab, Arnav Singhvi, Paridhi Maheshwari, Zhiyuan Zhang, Keshav Santhanam, Sri Vardhamanan, Saiful Haq, Ashutosh Sharma, Thomas~T. Joshi, Hanna Moazam, Heather Miller, Matei Zaharia, and Christopher Potts. 2023{\natexlab{b}}.
\newblock \href {https://arxiv.org/abs/2310.03714} {Dspy: Compiling declarative language model calls into self-improving pipelines}.
\newblock \emph{Preprint}, arXiv:2310.03714.

\bibitem[{Lewis et~al.(2020)Lewis, Perez, Piktus, Petroni, Karpukhin, Goyal, K\"{u}ttler, Lewis, Yih, Rockt\"{a}schel, Riedel, and Kiela}]{rag}
Patrick Lewis, Ethan Perez, Aleksandra Piktus, Fabio Petroni, Vladimir Karpukhin, Naman Goyal, Heinrich K\"{u}ttler, Mike Lewis, Wen-tau Yih, Tim Rockt\"{a}schel, Sebastian Riedel, and Douwe Kiela. 2020.
\newblock \href {https://proceedings.neurips.cc/paper_files/paper/2020/file/6b493230205f780e1bc26945df7481e5-Paper.pdf} {Retrieval-augmented generation for knowledge-intensive nlp tasks}.
\newblock In \emph{Advances in Neural Information Processing Systems}, volume~33, pages 9459--9474. Curran Associates, Inc.

\bibitem[{Lipscomb(2000)}]{Lipscomb2000-wk}
C~E Lipscomb. 2000.
\newblock Medical subject headings ({MeSH}).
\newblock \emph{Bulletin of the Medical Library Association}, 88(3):265--266.

\bibitem[{Liu et~al.(2021)Liu, Xu, Wen, Jiang, Wu, and Yuan}]{liu-etal-2021-temp}
Zichen Liu, Hongyuan Xu, Yanlong Wen, Ning Jiang, HaiYing Wu, and Xiaojie Yuan. 2021.
\newblock \href {https://doi.org/10.18653/v1/2021.emnlp-main.313} {{TEMP}: Taxonomy expansion with dynamic margin loss through taxonomy-paths}.
\newblock In \emph{Proceedings of the 2021 Conference on Empirical Methods in Natural Language Processing}, pages 3854--3863, Online and Punta Cana, Dominican Republic. Association for Computational Linguistics.

\bibitem[{Ma et~al.(2021)Ma, Chen, Wu, and Peng}]{ma-etal-2021-hyperexpan-taxonomy}
Mingyu~Derek Ma, Muhao Chen, Te-Lin Wu, and Nanyun Peng. 2021.
\newblock \href {https://doi.org/10.18653/v1/2021.findings-emnlp.353} {{H}yper{E}xpan: Taxonomy expansion with hyperbolic representation learning}.
\newblock In \emph{Findings of the Association for Computational Linguistics: EMNLP 2021}, pages 4182--4194, Punta Cana, Dominican Republic. Association for Computational Linguistics.

\bibitem[{Manzoor et~al.(2020)Manzoor, Li, Shrouty, and Leskovec}]{manzoor-2020}
Emaad Manzoor, Rui Li, Dhananjay Shrouty, and Jure Leskovec. 2020.
\newblock \href {https://doi.org/10.1145/3366423.3380271} {Expanding taxonomies with implicit edge semantics}.
\newblock In \emph{Proceedings of The Web Conference 2020}, WWW '20, page 2044–2054, New York, NY, USA. Association for Computing Machinery.

\bibitem[{Margiotta et~al.(2023)Margiotta, Croce, and Basili}]{margiotta_2023}
Daniele Margiotta, Danilo Croce, and Roberto Basili. 2023.
\newblock Taxosbert: Unsupervised taxonomy expansion through expressive semantic similarity.
\newblock In \emph{Deep Learning Theory and Applications}, pages 295--307, Cham. Springer Nature Switzerland.

\bibitem[{Milios et~al.(2023)Milios, Reddy, and Bahdanau}]{milios-etal-2023-context}
Aristides Milios, Siva Reddy, and Dzmitry Bahdanau. 2023.
\newblock \href {https://doi.org/10.18653/v1/2023.genbench-1.14} {In-context learning for text classification with many labels}.
\newblock In \emph{Proceedings of the 1st GenBench Workshop on (Benchmarking) Generalisation in NLP}, pages 173--184, Singapore. Association for Computational Linguistics.

\bibitem[{Reimers and Gurevych(2019)}]{reimers-gurevych-2019-sentence}
Nils Reimers and Iryna Gurevych. 2019.
\newblock \href {https://doi.org/10.18653/v1/D19-1410} {Sentence-{BERT}: Sentence embeddings using {S}iamese {BERT}-networks}.
\newblock In \emph{Proceedings of the 2019 Conference on Empirical Methods in Natural Language Processing and the 9th International Joint Conference on Natural Language Processing (EMNLP-IJCNLP)}, pages 3982--3992, Hong Kong, China. Association for Computational Linguistics.

\bibitem[{Shen et~al.(2020)Shen, Shen, Xiong, Wang, Wang, and Han}]{shen_2020}
Jiaming Shen, Zhihong Shen, Chenyan Xiong, Chi Wang, Kuansan Wang, and Jiawei Han. 2020.
\newblock \href {https://doi.org/10.1145/3366423.3380132} {Taxoexpan: Self-supervised taxonomy expansion with position-enhanced graph neural network}.
\newblock In \emph{Proceedings of The Web Conference 2020}, WWW '20, page 486–497, New York, NY, USA. Association for Computing Machinery.

\bibitem[{Shen et~al.(2018)Shen, Wu, Lei, Zhang, Ren, Vanni, Sadler, and Han}]{shen_2018}
Jiaming Shen, Zeqiu Wu, Dongming Lei, Chao Zhang, Xiang Ren, Michelle~T. Vanni, Brian~M. Sadler, and Jiawei Han. 2018.
\newblock \href {https://doi.org/10.1145/3219819.3220115} {Hiexpan: Task-guided taxonomy construction by hierarchical tree expansion}.
\newblock In \emph{Proceedings of the 24th ACM SIGKDD International Conference on Knowledge Discovery \& Data Mining}, KDD '18, page 2180–2189, New York, NY, USA. Association for Computing Machinery.

\bibitem[{Singhvi et~al.(2024)Singhvi, Shetty, Tan, Potts, Sen, Zaharia, and Khattab}]{singhvi2024}
Arnav Singhvi, Manish Shetty, Shangyin Tan, Christopher Potts, Koushik Sen, Matei Zaharia, and Omar Khattab. 2024.
\newblock \href {https://arxiv.org/abs/2312.13382} {Dspy assertions: Computational constraints for self-refining language model pipelines}.
\newblock \emph{Preprint}, arXiv:2312.13382.

\bibitem[{Wang et~al.(2022)Wang, Zhao, Zheng, and Liu}]{wang_2022}
Suyuchen Wang, Ruihui Zhao, Yefeng Zheng, and Bang Liu. 2022.
\newblock \href {https://doi.org/10.1145/3485447.3511943} {Qen: Applicable taxonomy completion via evaluating full taxonomic relations}.
\newblock In \emph{Proceedings of the ACM Web Conference 2022}, WWW '22, page 1008–1017, New York, NY, USA. Association for Computing Machinery.

\bibitem[{Wei et~al.(2022)Wei, Wang, Schuurmans, Bosma, ichter, Xia, Chi, Le, and Zhou}]{wei_2022}
Jason Wei, Xuezhi Wang, Dale Schuurmans, Maarten Bosma, brian ichter, Fei Xia, Ed~Chi, Quoc~V Le, and Denny Zhou. 2022.
\newblock \href {https://proceedings.neurips.cc/paper_files/paper/2022/file/9d5609613524ecf4f15af0f7b31abca4-Paper-Conference.pdf} {Chain-of-thought prompting elicits reasoning in large language models}.
\newblock In \emph{Advances in Neural Information Processing Systems}, volume~35, pages 24824--24837. Curran Associates, Inc.

\bibitem[{Wu and Palmer(1994)}]{wu-palmer-1994-verb}
Zhibiao Wu and Martha Palmer. 1994.
\newblock \href {https://doi.org/10.3115/981732.981751} {Verb semantics and lexical selection}.
\newblock In \emph{32nd Annual Meeting of the Association for Computational Linguistics}, pages 133--138, Las Cruces, New Mexico, USA. Association for Computational Linguistics.

\bibitem[{Wullschleger et~al.(2025)Wullschleger, Zarharan, Daly, Pouly, and Foster}]{wullschleger2025goldstandardproblemreferencefree}
Pascal Wullschleger, Majid Zarharan, Donnacha Daly, Marc Pouly, and Jennifer Foster. 2025.
\newblock \href {https://arxiv.org/abs/2505.11470} {No gold standard, no problem: Reference-free evaluation of taxonomies}.
\newblock \emph{Preprint}, arXiv:2505.11470.

\bibitem[{Xu et~al.(2023)Xu, Liu, Niu, Chen, Cai, Wen, and Yuan}]{xu-etal-2023-tacoprompt}
Hongyuan Xu, Ciyi Liu, Yuhang Niu, Yunong Chen, Xiangrui Cai, Yanlong Wen, and Xiaojie Yuan. 2023.
\newblock \href {https://doi.org/10.18653/v1/2023.emnlp-main.979} {{T}aco{P}rompt: A collaborative multi-task prompt learning method for self-supervised taxonomy completion}.
\newblock In \emph{Proceedings of the 2023 Conference on Empirical Methods in Natural Language Processing}, pages 15804--15817, Singapore. Association for Computational Linguistics.

\bibitem[{Yu et~al.(2020)Yu, Li, Shen, Feng, Sun, and Zhang}]{yu_2020}
Yue Yu, Yinghao Li, Jiaming Shen, Hao Feng, Jimeng Sun, and Chao Zhang. 2020.
\newblock \href {https://doi.org/10.1145/3394486.3403145} {Steam: Self-supervised taxonomy expansion with mini-paths}.
\newblock In \emph{Proceedings of the 26th ACM SIGKDD International Conference on Knowledge Discovery \& Data Mining}, KDD '20, page 1026–1035, New York, NY, USA. Association for Computing Machinery.

\bibitem[{Zeng et~al.(2021)Zeng, Lin, Yu, Cleland-Huang, and Jiang}]{zeng_2021}
Qingkai Zeng, Jinfeng Lin, Wenhao Yu, Jane Cleland-Huang, and Meng Jiang. 2021.
\newblock \href {https://doi.org/10.1145/3447548.3467308} {Enhancing taxonomy completion with concept generation via fusing relational representations}.
\newblock In \emph{Proceedings of the 27th ACM SIGKDD Conference on Knowledge Discovery \& Data Mining}, KDD '21, page 2104–2113, New York, NY, USA. Association for Computing Machinery.

\bibitem[{Zhang et~al.(2023)Zhang, Haddow, and Birch}]{pmlr-v202-zhang23m}
Biao Zhang, Barry Haddow, and Alexandra Birch. 2023.
\newblock \href {https://proceedings.mlr.press/v202/zhang23m.html} {Prompting large language model for machine translation: A case study}.
\newblock In \emph{Proceedings of the 40th International Conference on Machine Learning}, volume 202 of \emph{Proceedings of Machine Learning Research}, pages 41092--41110. PMLR.

\bibitem[{Zhang et~al.(2021)Zhang, Song, Zeng, Chen, Shen, Mao, and Li}]{Zhang_Song_Zeng_Chen_Shen_Mao_Li_2021}
Jieyu Zhang, Xiangchen Song, Ying Zeng, Jiaze Chen, Jiaming Shen, Yuning Mao, and Lei Li. 2021.
\newblock \href {https://doi.org/10.1609/aaai.v35i5.16596} {Taxonomy completion via triplet matching network}.
\newblock \emph{Proceedings of the AAAI Conference on Artificial Intelligence}, 35(5):4662--4670.

\end{thebibliography}
\clearpage

\appendix
\section{Implementation Details}

\subsection{Algorithms}
The proposed methods for completion and generation are formulated in more detail than  in the main section in algorithms \ref{alg:completion} and \ref{alg:generation} respectively.

\subsection{Embeddings}
\label{sec:embeddings}
For the retrieval step in our proposed models, we used FastText \cite{bojanowski-etal-2017-enriching}. In order to avoid a biased evaluation, we instead used sentence transformer embeddings\footnote{\url{https://huggingface.co/sentence-transformers/all-MiniLM-L6-v2}} \cite{reimers-gurevych-2019-sentence} for CSC to measure semantic similarity.

\subsection{NLI Verification}
\label{sec:nli_verify_app}
To minimise bias between model inference and evaluation, we use two different models. For the verification of generated concepts in the inference, we used \textit{ynie/roberta-large-snli\_mnli\_fever\_anli\_R1\_R2\_R3-nli}\footnote{\url{https://huggingface.co/ynie/roberta-large-snli\_mnli\_fever\_anli\_R1\_R2\_R3-nli}}, and for the NLIV metric during the evaluation \textit{facebook/bart-large-mnli}\footnote{\url{https://huggingface.co/facebook/bart-large-mnli}}.

\begin{algorithm*}
\caption{Taxonomy Completion}
\small\begin{algorithmic}[1]
\Require A query concept $q \in \mathcal{Q}$ to insert into taxonomy $\mathcal{T} = (\mathcal{E}, \mathcal{V})$ and a description $d_q$ for the query concept $q \in \mathcal{Q}$
\Ensure A set of predicted placements $\mathcal{Y}_q$ for the query concept $q$
\State $\mathcal{Y}_q \gets \emptyset$ \Comment{Set of predicted placements for the query $q$}
\State $R \gets \text{Retrieve}(q, \mathcal{T}, k)$ \Comment{Retrieve $k$ most relevant edges $R$ by cosine similarity to $q$}
\State $\mathcal{P} \gets \text{CoT}_p(q, R, d_q)$ \Comment{Generate candidate parent concepts using CoT prompting}
\State $\mathcal{P} \gets \{ p \in \mathcal{P} \mid \neg\text{contradicts}(\mathcal{d}_q, "\text{lemma}(q) \text{ is a } \text{lemma}(p)") \}$ \Comment{Validate parents with NLI}
\State $\mathcal{C} \gets \{ c \in \mathcal{V} \mid c \text{ is a child of any } p \in \mathcal{P} \}$ \Comment{Get candidate children}
\State $\mathcal{C} \gets \text{CoT}_c(q, \mathcal{C}, R, d_q)$ \Comment{Select valid children using CoT prompting}
\State $\mathcal{C} \gets \{ c \in \mathcal{C} \mid \text{entails}(\mathcal{d}_q, "\text{lemma}(c) \text{ is a } \text{lemma}(q)") \}$ \Comment{Validate children with NLI}
\For{each parent-child combination $(p, c) \in \mathcal{P} \times \mathcal{C}$}
    \If{$p \text{ is a parent of } c \text{ in } \mathcal{T}$}
        \State $\mathcal{Y}_q \gets \mathcal{Y}_q\cup \{(p, q, c)\}$ \Comment{Add valid placement to $\mathcal{Y}_q$}
    \EndIf
\EndFor
\end{algorithmic}
\label{alg:completion}
\end{algorithm*}

\begin{algorithm*}
\caption{Taxonomy Generation}
\small\begin{algorithmic}[1]
\Require A set of concepts $\mathcal{Q}$ to insert into taxonomy $\mathcal{\T} = (\mathcal{E}, \mathcal{V})$ and a description $d_q \in \mathcal{D}$ for each query concept $q \in \mathcal{Q}$
\Ensure A completed taxonomy $\mathcal{\T}$
\State $\mathcal{V} \gets \mathcal{Q}$
\State $\mathcal{E} \gets \emptyset$
\State $\mathcal{Q}_n \gets \{q_1, \dots, q_n\},\; q_i \overset {iid}\sim \text{Uniform}(\mathcal{Q})$ \Comment{Sample $n$ concepts from $\mathcal{Q}$}
\State $d_t \gets \text{CoT}_d(\mathcal{Q}_n)$ \Comment{Describe the potential taxonomy using CoT prompting}
\While{$|\mathcal{Q}| > 0$}
    \State $q \gets \text{Next}(\mathcal{Q})$  \Comment{Get next query $q$ from set of concepts to add}
    \State $R \gets \text{Retrieve}(q, \mathcal{T}, k)$ \Comment{Retrieve $k$ most relevant edges $R$ by cosine similarity to $q$}
    \State $\mathcal{P} \gets \text{CoT}_p(q, R, d_q, d_t)$ \Comment{Generate candidate parent concepts using CoT prompting}
    \State $\mathcal{P} \gets \{ p \in \mathcal{P}_q \mid \neg\text{contradicts}(d_q, "\text{lemma}(q) \text{ is a } \text{lemma}(p)") \}$ \Comment{Validate parents with NLI}
    \State $\mathcal{C} \gets \{ c \in \mathcal{V} \mid c \text{ is a child of any } p \in \mathcal{P} \}$ \Comment{Get candidate children}
    \State $\mathcal{C} \gets \text{CoT}_c(q, \mathcal{C}, R, d_q, d_t)$ \Comment{Select valid children using CoT prompting}
    \State $\mathcal{C} \gets \{ c \in \mathcal{C} \mid \text{entails}(d_q, "\text{lemma}(c) \text{ is a } \text{lemma}(q)") \}$ \Comment{Validate children with NLI}
    \State $\mathcal{N} \gets \mathcal{P} \setminus \mathcal{V}$ \Comment{Get newly generated concepts}
    \State $\mathcal{Q} \gets \mathcal{Q} \cup \mathcal{N}$ \Comment{Update set of concepts to add}
    \State $\T \gets \text{InsertParents}(q, \mathcal{P}, \mathcal{\T})$ \Comment{Insert new parent-query edges into taxonomy.}
    \State $\T \gets \text{InsertChildren}(q, \mathcal{C}, \T)$ \Comment{Insert new query-child edges into taxonomy.}
    \State $\mathcal{Q} \gets \mathcal{Q} \setminus \{q\} $ \Comment{Remove added concept}
\EndWhile
\end{algorithmic}
\label{alg:generation}
\end{algorithm*}

\subsection{DSPy}
\label{sec:dspy}
The \texttt{DSPy} library \cite{khattab2023dspycompilingdeclarativelanguage}, enables us to use \gls{RAG} in an off-the-shelf manner and to tune prompts per model and datasets 
with a hyper-parameter-tuning like approach. 

Due to issues we encountered with Llama3 and the \texttt{DSPy} library, we customized the template \texttt{DSPy} uses to generate prompts, by more clearly separating the input and output fields\footnote{\url{https://github.com/wullli/foodtaxo}}. The customization is apparent in our examples of prompts and outputs, as shown in Section~\ref{sec:prompts}.

\texttt{DSPy} provides optimizers which can be used to tune prompts given validation and training data. We evaluated the automated tuning of instruction texts with their \texttt{COPRO} optimizer. This optimizer generates variations of a predefined prompt using a language model and evaluates its effectiveness on validation examples. It keeps the most promising examples and generates further variations them. Results of the comparison of instruction-tuned (COPRO) against our handcrafted prompts can be found in Table~\ref{tab:eval_tuned}. We randomly sampled 20 concepts from our validation and training sets respectively and ran the optimizer by generating 5 initial variations of our default prompt and allowing 2 subsequent variations on each.

\begin{table*}
    \centering
    \small
    \resizebox{1\textwidth}{!}
    {
    \begin{tabular}{llrrrrrrrrrrrrrr}
        \toprule
        \multirow{2}{*}{\textbf{Dataset}} & 
        \multirow{2}{*}{\textbf{Model}} &
        \multicolumn{4}{c}{\textbf{Total}} && 
        \multicolumn{4}{c}{\textbf{Non-Leaf}} && 
        \multicolumn{4}{c}{\textbf{Leaf}}\\
        \cmidrule{3-6}
        \cmidrule{8-11}
        \cmidrule{13-16}
        && WPS & F1 & P & R && WPS & F1 & P & R && WPS & F1 & P & R\\
        \midrule
        \multirow{2}{*}{SemEval-Food}
        & Llama-3 Zero-Shot & \textbf{0.8177} & \underline{0.2192} & \underline{0.3780} & \underline{0.1543} &  & \textbf{0.8050} & \underline{0.0508} & \underline{0.2381} & \underline{0.0284} &  & \textbf{0.8203} & \underline{0.3568} & \underline{0.4057} & \underline{0.3185} \\
        & Llama-3 Zero-Shot* & 0.7723 & \textbf{0.2367} & \textbf{0.4250} & \textbf{0.1640} &  & \underline{0.7407} & \textbf{0.0619} & \textbf{0.3333} & \textbf{0.0341} &  & \underline{0.7788} & \textbf{0.3797} & \textbf{0.4412} & \textbf{0.3333} \\
        
        \midrule
        
        \multirow{2}{*}{MeSH}
        & Llama-3 Zero-Shot & \textbf{0.8549} & \textbf{0.1662} & \underline{0.2877} & \textbf{0.1169} &  & \textbf{0.8645} & \textbf{0.0845} & \underline{0.2460} & \textbf{0.0510} &  & \textbf{0.8504} & \textbf{0.2597} & \textbf{0.3071} & \textbf{0.2250} \\
        & Llama-3 Zero-Shot* & 0.8397 & \underline{0.1610} & \textbf{0.2882} & \underline{0.1117} &  & 0.8473 & \underline{0.0824} & \textbf{0.2500} & \underline{0.0493} &  & 0.8361 & \underline{0.2518} & \underline{0.3059} & \underline{0.2140} \\
        \bottomrule
    \end{tabular}
    }
  \caption{Comparison of instruction tuning using \texttt{DSPy} optimizers. All scores that are not significantly different to the best model according to a two-sided paired randomization test ($\alpha = 0.05$) with 1k resamples are underlined. Models marked with an asterisk (*) were instruction tuned using \texttt{DSPy}.}
  \label{tab:eval_tuned} 
\end{table*}

\subsection{Processing LLM Outputs}
It is possible, that an LLM predicts a set of parents for a concept, where inside that set one parent is already an ancestor of another in the taxonomy. In such a case, we select the most specific concept (furthest from the root).

\subsection{Concept Descriptions}
\label{sec:concept_desc}
The MeSH, SemEval-Food and SemEval-Verb datasets include descriptions for all concepts. For Wikidata and CookBook we have no concept descriptions and instead generated descriptions using \texttt{gpt-4o-mini}.

\subsection{Evaluation metrics}
We notice that some test concepts in SemEval-Verb do not have gold standard positions. We do not calculate any scores for such concepts but average over the available gold standards. Note that, since we follow \citet{Zhang_Song_Zeng_Chen_Shen_Mao_Li_2021} and assume that the task is $N$ independent attachment problems, it is possible that we create cycles by inserting all predicted placements into an existing taxonomy. The calculation of quality attributes, such as robustness, requires the insertion of concepts to calculate scores. In such cases, we simply drop placements that would lead to cycles and do not consider them during the calculation. The standard metrics used in completion are described below. Note that for a position to be considered correct, both parent and child of the query concept need to be correct. A correctly predicted parent with an incorrectly predicted child will result in a false positive and vice versa.

\paragraph{Recall (R)} How many of the true positions were correctly predicted by the model.
\begin{equation}
\frac{TP}{TP+FN}
\end{equation}
\paragraph{Precision (P)} How many of the predicted positions were correct.
\begin{equation}
\frac{TP}{TP+FP}
\end{equation}
\paragraph{F1-score (F1)} The harmonic mean of the precision and recall for the positions.
\begin{equation}
2 \cdot \frac{P\cdot R}{P + R}
\label{eq:f1}
\end{equation}

\section{Experiment Details}
We reused implementations for the baselines from \citet{xu-etal-2023-tacoprompt} and adjusted them for our setting by adding the functionality to output the best placements (triplets) for a query instead of a ranked list, so that we could subsequently calculate F1, precision, and recall. We ensured the quality of the implementation of our metrics by validating them against metrics used by \citet{xu-etal-2023-tacoprompt}.

\subsection{Baselines} 
\label{sec:baselines}
We utilized the following state-of-the-art taxonomy completion techniques as baselines for comparison with our proposed method. 

\begin{itemize}
    \item \textbf{Arborist}: \citet{manzoor-2020} propose Arborist, an approach to expand textual taxonomies by predicting parents of new nodes with unobserved heterogeneous edge semantics. Arborist learns latent edge representations and node embeddings, optimizing a large-margin ranking loss to minimize the shortest-path distance between predicted and actual parents.

    \item \textbf{QEN}: \citet{wang_2022} propose the Quadruple Evaluation Network (QEN), a taxonomy completion framework using term descriptions, pre-trained language models, and code attention for accurate inference while reducing computation. QEN evaluates parent-child and sibling relations to enhance accuracy and reduce noise from pseudo-leaves.

    \item \textbf{TEMP}: \citet{liu-etal-2021-temp} present TEMP, a self-supervised taxonomy expansion method that predicts new concept positions by ranking generated paths. TEMP utilizes pre-trained contextual encoders for taxonomy construction and hypernym detection. \citet{liu-etal-2021-temp} show that pre-trained contextual embeddings capture hypernym-hyponym relations effectively.
    
    \item \textbf{TMN}: \citet{Zhang_Song_Zeng_Chen_Shen_Mao_Li_2021} introduce "taxonomy completion" and propose the Triplet Matching Network (TMN) to find hypernym and hyponym concepts for a query. TMN, featuring a primal scorer, auxiliary scorers, and a channel-wise gating mechanism, outperforms existing methods. 
    
    \item \textbf{TacoPrompt}: \citet{xu-etal-2023-tacoprompt} introduce TacoPrompt, employing triplet semantic matching via prompt learning to address imbalanced data, a contextual approach to connect subtask results with final predictions. TacoPrompt also leverages a two-stage retrieval and re-ranking method to enhance inference efficiency.
    
    \item \textbf{TaxoExpan}: \citet{shen_2020} present TaxoExpan, a self-supervised framework for expanding taxonomies by automatically generating ⟨query concept, anchor concept⟩ pairs from existing taxonomies. TaxoExpan uses this data to predict whether a query concept is the direct hyponym of an anchor concept.
    
\end{itemize}

\setlength\arrayrulewidth{1pt}
\begin{table}[ht!]
    \centering
    \small
    \resizebox{0.48\textwidth}{!}
    {
    \setlength{\tabcolsep}{6pt}
    {
    \begin{tabular}{lrrrrrr}
        \toprule
         \textbf{Dataset} & $|\mathbfcal{V}|$ & $|\mathbfcal{E}|$& \textbf{$\mathbf{D}$} & \textbf{$\mathbf{|L|}$} & \textbf{$\mathbf{\frac{|L|}{|\mathbfcal{V}|}}$} & $\mathbf{B}$\\ 
         \midrule
        SemEval-Food & 1486 & 1576 & 9 & 1184 & 0.80 & 5.08 \\
        SemEval-Verb & 13936 & 13407 & 13 & 10360 & 0.74 & 4.12 \\
        MeSH & 9710 & 10496 & 11 & 5502 & 0.57 & 3.88 \\
        Wikitax & 941 & 973 & 7 & 754 & 0.80 & 5.20 \\
        CookBook & 1985 & 1984 & 4 & 1795 & 0.90 & 10.44 \\
        \midrule        
        Generated Recipe1M & 12376 & 12745 & 15 & 10156 & 0.82 & 5.74 \\
        Generated MeSH & 6908 & 6858 & 10 & 5712 & 0.83 & 5.65 \\
        Generated SemEval-Food & 1213 & 1257 & 11 & 1130 & 0.93 & 15.14 \\
        \midrule   
        Generated SemEval-Food (w/o NLI) & 1203 & 1216 & 6 & 1122 & 0.93 & 15.01 \\
        Generated SemEval-Food (w/o Backtracking) & 1228 & 1272 & 7 & 1108 & 0.90 & 10.60 \\
        Generated SemEval-Food (w/o Generation) & 1233 & 1251 & 12 & 1135 & 0.92 & 12.77 \\
        \bottomrule
    \end{tabular}}
    }
  \caption{Statistics regarding generated and brenchmark taxonomies. $|\mathcal{V}|$, $|\mathcal{E}|$, $D$, $|L|$, $\frac{|L|}{|\mathcal{V}|}$, $B$ represent the node number, edge number, depth, the number of leaves, the ratio of leaves and the branching factor of the taxonomy.}
  \label{tab:tax_stats_appendix} 
\end{table}

\subsection{Prompt} 
\label{sec:prompts_app}
We show the default handcrafted prompts for generating parent (Listing~\ref{lst:parent_prompt}) and child concepts (Listing~\ref{lst:child_prompt}) of a query. We optimized these prompts by manual trial and error on the validation data. For spacing reasons, we display only two lines of context for both prompts and remove double new-lines. Note that the prompts for generation are slightly different. 

In this section we provide examples of model prompts for DSPy-tuned and untuned models. The same example is provided for each setting. The context is only displayed for the first Parent and Child generation prompt to limit verbosity.

\subsubsection{Completion: Tuned Zero-Shot Parents}

\begin{lstlisting}[linewidth=\linewidth,language=prompt]
// Prompt
Which are the most specific parent concepts of the given child concept in a taxonomy considering the context?
In your reasoning, state how the parent concepts are a supertype of the child concept.
Do not add additional comments or information, only return the output in the described format.

---

Input description.

Context: List of existing parent-child (supertype-subtype) relations in the taxonomy.

Child: Child concept (subtype) that you need to place in a taxonomy.

Description: Description of the child concept.

---

Follow the following format.

Reasoning: Let's think step by step in order to ${produce the parents}. We ...

Interpretation: Description of the child concept in relation to the context taxonomy. Infer what is meant by the child concept from the context.

Parents: Comma separated list of one or more parents of the child concept. Valid parents are in the context. If there are no suitable parents, return None.

---

Context:
```
...
```

Child: sweetening

Description: sweetening is something added to foods to make them taste sweeter

Reasoning: Let's think step by step in order to
\end{lstlisting}

\begin{lstlisting}[linewidth=\linewidth,language=prompt,style=generated]
// Output
Reasoning: Let's think step by step in order to find the parents of the child concept "sweetening". We can infer that sweetening is related to making food taste sweeter, which is closely related to sugar and other sweet substances.

Interpretation: The child concept "sweetening" refers to an additive that enhances the sweetness of food, which is a common theme in the context taxonomy.

Parents: sugar, syrup
\end{lstlisting}

\subsubsection{Completion: Tuned Zero-Shot Children}

\begin{lstlisting}[linewidth=\linewidth,language=prompt]
// Prompt
Which of the candidates are child concepts (subtypes) of the given parent concept (supertype) in a taxonomy?
The context shows existing parents and children concepts and whether the children are leaves.
In your reasoning, state how the parent concept is a supertype of the selected child concepts.
Do not add additional comments or information, only return the output in the described format.

---

Input description.

Context: List of existing parent-child (supertype-subtype) relations in the taxonomy.

Candidates: Candidate children of the concept separated by commas to select from.

Parent: Parent concept that you need to place in a taxonomy.

Description: Description of the parent concept.

Interpretation: Description of the child concept in relation to the taxonomy.

---

Follow the following format.

Reasoning: Let's think step by step in order to ${produce the children}. We ...

Leaf: Whether the parent concept should be added as a leaf (has no children). Answer with Yes or No.

Children: Comma separated list of candidates that are children of the parent concept in a taxonomy.A child concept must be a type of the parent concept.Separate with commas.

---

Context:
```
...
```

Candidates: corn syrup, molasses, maple syrup, grenadine, sorghum, sugar syrup, corn sugar, lump sugar, caramel, granulated sugar, sugarloaf, beet sugar, brown sugar

Parent: sweetening

Description: sweetening is something added to foods to make them taste sweeter

Interpretation: The child concept "sweetening" refers to an additive that enhances the sweetness of food, which is a common theme in the context taxonomy.

Reasoning: Let's think step by step in order to
\end{lstlisting}

\begin{lstlisting}[linewidth=\linewidth,language=prompt,style=generated]
// Output
Reasoning: Let's think step by step in order to identify the children of the concept "sweetening". We can see that the context taxonomy includes various types of sweet substances and additives, which are used to make food taste sweeter. Based on this, we can identify the child concepts that are types of sweetening agents.

Leaf: No

Children: sugar syrup, corn syrup, molasses, maple syrup, grenadine, sorghum, caramel, granulated sugar, sugarloaf, beet sugar, brown sugar
\end{lstlisting}

\subsubsection{Completion: Few-Shot Parents}
In this example, we can see that backtracking (retrying with instructions) was applied, since the initial answer failed to fulfill the required constraints. Remarks prefixed with double slashes (/) are not actually part of the raw prompt, but serve as comments in the illustrations that follow.

\begin{lstlisting}[linewidth=\linewidth,language=prompt]
// Prompt
Which are the most specific parent concepts of the given child concept in a taxonomy considering the context?
In your reasoning, state how the parent concepts are a supertype of the child concept.
Do not add additional comments or information, only return the output in the described format.

---

Input description.

Context: List of existing parent-child (supertype-subtype) relations in the taxonomy.

Child: Child concept (subtype) that you need to place in a taxonomy.

Description: Description of the child concept.

Previous Reasoning: past Reasoning: with errors

Previous Interpretation: past Interpretation: with errors

Previous Parents: past Parents: with errors

Instructions: Some instructions you must satisfy

---

Follow the following format.

Reasoning: Let's think step by step in order to ${produce the parents}. We ...

Interpretation: Description of the child concept in relation to the context taxonomy. Infer what is meant by the child concept from the context.

Parents: Comma separated list of one or more parents of the child concept. Valid parents are in the context. If there are no suitable parents, return None.

---

Context:
```tetrazzini, chicken tetrazzini
dish, chicken provencale
dish, chicken marengo
casserole, chicken casserole
broth, chicken broth```

Child: chicken paprika

Description: chicken paprika is chicken simmered in broth with onions and paprika then mixed with sour cream

Reasoning: Let's think step by step in order to Reasoning: Let's think step by step in order to find the parents of "chicken paprika". We can see that "chicken paprika" is a type of dish, as it is a preparation of chicken with certain ingredients. Therefore, it can be classified under "dish".

Interpretation: "chicken paprika" is a type of dish that involves simmering chicken in broth with onions and paprika, and then mixing it with sour cream.

Parents: dish

---

Context:
```powdered sugar, icing sugar
flavorer, sugar
syrup, sugar syrup
sugar, brown sugar
sugar, beet sugar```

Child: granulated sugar

Description: granulated sugar is sugar in the form of small grains

Reasoning: Let's think step by step in order to Reasoning: Let's think step by step in order to find the parents of granulated sugar. We can see that granulated sugar is a type of sugar, so sugar is a supertype of granulated sugar.

Interpretation: Granulated sugar is a type of sugar that is in the form of small grains, which is a specific characteristic of this type of sugar.

Parents: sugar

---

Context:
```wine, blush wine
wine, canary wine
beverage, wine
vinegar, wine vinegar
sauce, wine sauce```

Child: red wine

Description: red wine is wine having a red color derived from skins of dark-colored grapes

Reasoning: Let's think step by step in order to Reasoning: Let's think step by step in order to find the parents of "red wine". We know that "red wine" is a type of "wine" because it is described as "wine having a red color derived from skins of dark-colored grapes".

Interpretation: "Red wine" is a specific type of wine, which is a beverage.

Parents: wine, beverage

---

Context:
```tea, ice tea
tea, green tea
tea, herb tea
herb, tea
tisane, camomile tea```

Child: sun tea

Description: sun tea is tea made by exposing tea leaves steeped in water to the direct rays of the sun; usually served with ice

Reasoning: Let's think step by step in order to Reasoning: Let's think step by step in order to find the parents of sun tea. We can see that sun tea is a type of tea, as it is made from tea leaves steeped in water.

Interpretation: Sun tea is a type of tea that is made using a specific method, which is by exposing the tea leaves to the direct rays of the sun.

Parents: tea

---

Context:
```
...
```

Child: sweetening

Description: sweetening is something added to foods to make them taste sweeter

Previous Reasoning: Reasoning: Let's think step by step in order to find the parents of "sweetening". We can see that "sweetening" is a type of "foodstuff" or an ingredient added to food, but there is no direct match in the context. However, we can relate it to "sweet" which is a type of food characteristic.

Previous Interpretation: "Sweetening" is an ingredient or substance added to food to make it sweeter, which is a characteristic of some foods.

Previous Parents: None

Instructions: None is not a valid parent.

Reasoning: Let's think step by step in order to
\end{lstlisting}

\begin{lstlisting}[linewidth=\linewidth,language=prompt,style=generated]
// Output
Reasoning: Let's think step by step in order to find the parents of "sweetening". We can see that "sweetening" is something added to foods to make them taste sweeter, which is related to the concept of "sweet" as a characteristic of food. Although there is no direct match in the context, we can relate it to "foodstuff" as it is an ingredient added to food.

Interpretation: "Sweetening" is an ingredient or substance added to food to make it sweeter, which is a characteristic of some foods.

Parents: foodstuff
\end{lstlisting}

\subsubsection{Completion: Few-Shot Children}
\begin{lstlisting}[linewidth=\linewidth,language=prompt]
// Prompt
Which of the candidates are child concepts (subtypes) of the given parent concept (supertype) in a taxonomy?
The context shows existing parents and children concepts and whether the children are leaves.
In your reasoning, state how the parent concept is a supertype of the selected child concepts.
Do not add additional comments or information, only return the output in the described format.

---

Input description.

Context: List of existing parent-child (supertype-subtype) relations in the taxonomy.

Candidates: Candidate children of the concept separated by commas to select from.

Parent: Parent concept that you need to place in a taxonomy.

Description: Description of the parent concept.

Interpretation: Description of the child concept in relation to the taxonomy.

---

Follow the following format.

Reasoning: Let's think step by step in order to ${produce the children}. We ...

Leaf: Whether the parent concept should be added as a leaf (has no children). Answer with Yes or No.

Children: Comma separated list of candidates that are children of the parent concept in a taxonomy.A child concept must be a type of the parent concept.Separate with commas.

---

Context:
```tetrazzini (Non-Leaf), chicken tetrazzini (Leaf)
dish (Non-Leaf), chicken provencale (Leaf)
dish (Non-Leaf), chicken marengo (Leaf)
casserole (Non-Leaf), chicken casserole (Leaf)
broth (Non-Leaf), chicken broth (Leaf)```

Candidates: chicken cordon bleu, croquette, pudding, pasta, succotash, chow mein, cottage pie, spaghetti and meatballs, poi, jambalaya, roulade, swiss steak, tamale pie, bacon and eggs, enchilada, barbecue, meat loaf, patty, lobster thermidor, potpie, coquilles saint jacques, sauerbraten, coq au vin, sauerkraut, tetrazzini, moussaka, refried beans, fondue, dolmas, steak au poivre, viand, sukiyaki, timbale, porridge, scallopine, seafood newburg, lutefisk, frittata, omelet, soup, pepper steak, spanish rice, galantine, barbecued wing, salisbury steak, sashimi, couscous, fish and chips, piece de resistance, veal cordon bleu, hash, chop suey, scotch egg, scrambled eggs, poached egg, chicken marengo, casserole, terrine, macedoine, pizza, meatball, welsh rarebit, osso buco, kishke, chicken paprika, carbonnade flamande, shirred egg, scampi, mold, taco, pork and beans, bitok, french toast, burrito, scrapple, haggis, pheasant under glass, maryland chicken, beef bourguignonne, boiled dinner, rijsttaffel, chicken and rice, schnitzel, kabob, beef wellington, risotto, paella, tempura, special, souffle, mousse, fish stick, tostada, frog legs, chili, snack food, ramekin, ham and eggs, boiled egg, chicken provencale, rissole, pilaf, applesauce, moo goo gai pan, kedgeree, stew, coleslaw, tossed salad, herring salad, crab louis, tuna fish salad, molded salad, pasta salad, potato salad, salmagundi, tabbouleh, chicken salad, fruit salad, lamb curry, raita, fruit custard, bavarian cream, creme caramel, knish, pirogi, apple turnover, samosa

Parent: chicken paprika

Description: chicken paprika is chicken simmered in broth with onions and paprika then mixed with sour cream

Interpretation: "chicken paprika" is a type of dish that involves simmering chicken in broth with onions and paprika, and then mixing it with sour cream.

Reasoning: Let's think step by step in order to Reasoning: Let's think step by step in order to identify the child concepts of "chicken paprika". We can see that "chicken paprika" is a type of dish, and it has a specific preparation method and ingredients. Therefore, it is a subtype of "dish" and cannot have children from the given candidates.

Leaf: Yes

Children: Reasoning: Let's think step by step in order to identify the child concepts of "chicken paprika". We can see that "chicken paprika" is a type of dish, and it has a specific preparation method and ingredients. Therefore, it is a subtype of "dish" and cannot have children from the given candidates. Leaf: Yes Children:

---

Context:
```powdered sugar (Non-Leaf), icing sugar (Leaf)
flavorer (Non-Leaf), sugar (Non-Leaf)
syrup (Non-Leaf), sugar syrup (Leaf)
sugar (Non-Leaf), brown sugar (Leaf)
sugar (Non-Leaf), beet sugar (Leaf)```

Candidates: corn sugar, lump sugar, caramel, granulated sugar, sugarloaf, beet sugar, brown sugar

Parent: granulated sugar

Description: granulated sugar is sugar in the form of small grains

Interpretation: Granulated sugar is a type of sugar that is in the form of small grains, which is a specific characteristic of this type of sugar.

Reasoning: Let's think step by step in order to Reasoning: Let's think step by step in order to identify the children of granulated sugar. Since granulated sugar is a type of sugar, we can look for candidates that are more specific types of granulated sugar.

Leaf: No

Children: sugarloaf

---

Context:
```wine (Non-Leaf), blush wine (Leaf)
wine (Non-Leaf), canary wine (Leaf)
beverage (Non-Leaf), wine (Non-Leaf)
vinegar (Non-Leaf), wine vinegar (Leaf)
sauce (Non-Leaf), wine sauce (Leaf)```

Candidates: varietal, burgundy, table wine, dubonnet, retsina, vintage, cotes de provence, mulled wine, blush wine, jug wine, vermouth, fortified wine, bordeaux, plonk, altar wine, tokay, red wine, riesling, sauterne, liebfraumilch, muscadet, pinot blanc, vouvray, verdicchio, sauvignon blanc, chardonnay, canary wine, yquem, straw wine, cold duck, champagne

Parent: red wine

Description: red wine is wine having a red color derived from skins of dark-colored grapes

Interpretation: "Red wine" is a specific type of wine, which is a beverage.

Reasoning: Let's think step by step in order to Reasoning: Let's think step by step in order to determine the children of "red wine". Since "red wine" is a type of wine, we can look for candidates that are specific types of red wine.

Leaf: No

Children: burgundy, bordeaux, tokay, riesling, pinot blanc, cold duck

---

Context:
```tea (Non-Leaf), ice tea (Leaf)
tea (Non-Leaf), green tea (Non-Leaf)
tea (Non-Leaf), herb tea (Non-Leaf)
herb (Non-Leaf), tea (Non-Leaf)
tisane (Non-Leaf), camomile tea (Leaf)```

Candidates: ice tea, herb tea, cuppa, cambric tea, sun tea, oolong, tea bag, black tea, green tea

Parent: sun tea

Description: sun tea is tea made by exposing tea leaves steeped in water to the direct rays of the sun; usually served with ice

Interpretation: Sun tea is a type of tea that is made using a specific method, which is by exposing the tea leaves to the direct rays of the sun.

Reasoning: Let's think step by step in order to Reasoning: Let's think step by step in order to determine the children of sun tea. Since sun tea is a type of tea, we can look for candidates that are also types of tea.

Leaf: No

Children: herb tea, green tea, black tea, ice tea

---

Context:
```
...
```

Candidates: concoction, frozen food, canned food, coloring, dehydrated food, concentrate, meal, starches, juice, dairy product, tapioca, flour, carob, breadstuff, egg, ingredient, perishable, roughage, corn, pearl barley, oat, wild rice, grist, malt, buckwheat, cracked wheat, millet, groats, criollo

Parent: sweetening

Description: sweetening is something added to foods to make them taste sweeter

Interpretation: "Sweetening" is an ingredient or substance added to food to make it sweeter, which is a characteristic of some foods.

Reasoning: Let's think step by step in order to
\end{lstlisting}

\begin{lstlisting}[language=prompt,style=generated]
// Output
Reasoning: Let's think step by step in order to determine the children of "sweetening". Since "sweetening" is an ingredient or substance added to food to make it sweeter, we can look for candidates that are specific types of sweetening agents.

Leaf: No

Children: tapioca, carob
\end{lstlisting}

\subsubsection{Completion: Tuned Few-Shot Parents}

\begin{lstlisting}[linewidth=\linewidth,language=prompt]
// Prompt
Which are the most specific parent concepts of the given child concept in a taxonomy considering the context?
In your reasoning, state how the parent concepts are a supertype of the child concept.
Do not add additional comments or information, only return the output in the described format.

---

Input description.

Context: List of existing parent-child (supertype-subtype) relations in the taxonomy.

Child: Child concept (subtype) that you need to place in a taxonomy.

Description: Description of the child concept.

---

Follow the following format.

Reasoning: Let's think step by step in order to ${produce the parents}. We ...

Interpretation: Description of the child concept in relation to the context taxonomy. Infer what is meant by the child concept from the context.

Parents: Comma separated list of one or more parents of the child concept. Valid parents are in the context. If there are no suitable parents, return None.

---

Context:
```foodstuff, flour
meal, pea flour
starches, cracked wheat bread
nutriment, wheat germ
starches, gluten bread```

Child: wheat flour

Description: wheat flour is flour prepared from wheat

Reasoning: Let's think step by step in order to Reasoning: Let's think step by step in order to find the parents of wheat flour. We know that wheat flour is a type of flour, and flour is a foodstuff. Therefore, foodstuff is a parent of wheat flour. Additionally, since wheat flour is a specific type of flour, flour is also a parent.

Interpretation: Wheat flour is a type of flour prepared from wheat, which is a foodstuff.

Parents: foodstuff, flour

---

Context:
```sauce, hard sauce
candy, hard candy
hard candy, barley sugar
cider, sweet cider
hard candy, sourball```

Child: hard cider

Description: hard cider is alcoholic drink from fermented cider; `cider' and `cyder' are European (especially British) usages for the fermented beverage

Reasoning: Let's think step by step in order to Reasoning: Let's think step by step in order to find the parents of "hard cider". We can see that "hard cider" is a type of "cider" because it is a specific type of cider that is fermented.

Interpretation: "hard cider" is a specific type of cider that is fermented, which makes it an alcoholic drink.

Parents: cider

---

Context:
```sauce, plum sauce
pudding, pease pudding
dessert, pudding
dish, pudding
pudding, carrot pudding```

Child: plum pudding

Description: plum pudding is a rich steamed or boiled pudding that resembles cake

Reasoning: Let's think step by step in order to Reasoning: Let's think step by step in order to find the parents of plum pudding. We can see that plum pudding is a type of pudding, and pudding is a type of dish and dessert. Also, plum sauce is a sauce, but it's not directly related to plum pudding.

Interpretation: Plum pudding is a specific type of pudding that is a dish and a dessert.

Parents: pudding, dish, dessert

---

Context:
```dish, cottage pie
dessert, apple dumpling
quick bread, corn cake
sweet pickle, bread and butter pickle
conserve, apple butter```

Child: tamale pie

Description: tamale pie is a meat mixture covered with cornbread topping that resembles a Mexican dish

Reasoning: Let's think step by step in order to Reasoning: Let's think step by step in order to find the parents of tamale pie. We can see that tamale pie is a type of dish, and it resembles a Mexican dish, so it can be classified under dish.

Interpretation: Tamale pie is a type of dish that has a cornbread topping, which is a characteristic of quick bread.

Parents: dish, quick bread

---

Context:
```
...
```

Child: sweetening

Description: sweetening is something added to foods to make them taste sweeter

Reasoning: Let's think step by step in order to
\end{lstlisting}

\begin{lstlisting}[linewidth=\linewidth,language=prompt,style=generated]
Reasoning: Let's think step by step in order to find the parents of sweetening. We can see that sweetening is related to sugar, as it is something added to foods to make them taste sweeter. Sugar is a type of nutriment, so nutriment can be a parent of sweetening.

Interpretation: Sweetening is something added to foods to make them taste sweeter, which is related to sugar and nutriment.

Parents: sugar, nutriment
\end{lstlisting}

\subsubsection{Completion: Tuned Few-Shot Children}

\begin{lstlisting}[linewidth=\linewidth,language=prompt]
// Prompt
Which of the candidates are child concepts (subtypes) of the given parent concept (supertype) in a taxonomy?
The context shows existing parents and children concepts and whether the children are leaves.
In your reasoning, state how the parent concept is a supertype of the selected child concepts.
Do not add additional comments or information, only return the output in the described format.

---

Input description.

Context: List of existing parent-child (supertype-subtype) relations in the taxonomy.

Candidates: Candidate children of the concept separated by commas to select from.

Parent: Parent concept that you need to place in a taxonomy.

Description: Description of the parent concept.

Interpretation: Description of the child concept in relation to the taxonomy.

---

Follow the following format.

Reasoning: Let's think step by step in order to ${produce the children}. We ...

Leaf: Whether the parent concept should be added as a leaf (has no children). Answer with Yes or No.

Children: Comma separated list of candidates that are children of the parent concept in a taxonomy.A child concept must be a type of the parent concept.Separate with commas.

---

Context:
```foodstuff (Non-Leaf), flour (Non-Leaf)
meal (Non-Leaf), pea flour (Leaf)
starches (Non-Leaf), cracked wheat bread (Leaf)
nutriment (Non-Leaf), wheat germ (Leaf)
starches (Non-Leaf), gluten bread (Leaf)```

Candidates: soybean meal, semolina, wheat flour, plain flour

Parent: wheat flour

Description: wheat flour is flour prepared from wheat

Interpretation: Wheat flour is a type of flour that is prepared from wheat, which is a foodstuff.

Reasoning: Let's think step by step in order to Reasoning: We can see that wheat flour is a type of flour, and semolina is also a type of flour. Therefore, semolina is a type of wheat flour. Similarly, plain flour is also a type of wheat flour.

Leaf: No

Children: semolina, plain flour

---

Context:
```fricassee (Non-Leaf), chicken stew (Leaf)
goulash (Non-Leaf), pork and veal goulash (Leaf)
mousse (Non-Leaf), chicken mousse (Leaf)
dish (Non-Leaf), chicken and rice (Non-Leaf)
stew (Non-Leaf), beef stew (Leaf)```

Candidates: chicken cordon bleu, croquette, pudding, pasta, succotash, chow mein, cottage pie, spaghetti and meatballs, poi, jambalaya, roulade, swiss steak, tamale pie, bacon and eggs, enchilada, barbecue, meat loaf, patty, lobster thermidor, potpie, coquilles saint jacques, sauerbraten, coq au vin, sauerkraut, tetrazzini, moussaka, refried beans, fondue, dolmas, steak au poivre, viand, sukiyaki, timbale, porridge, scallopine, seafood newburg, lutefisk, frittata, omelet, soup, pepper steak, spanish rice, galantine, barbecued wing, salisbury steak, sashimi, couscous, fish and chips, piece de resistance, veal cordon bleu, hash, chop suey, scotch egg, scrambled eggs, poached egg, chicken marengo, casserole, terrine, macedoine, pizza, meatball, welsh rarebit, osso buco, kishke, chicken paprika, carbonnade flamande, shirred egg, scampi, mold, taco, pork and beans, bitok, french toast, burrito, scrapple, haggis, pheasant under glass, maryland chicken, beef bourguignonne, boiled dinner, rijsttaffel, chicken and rice, schnitzel, kabob, beef wellington, risotto, paella, tempura, special, souffle, mousse, fish stick, tostada, frog legs, chili, snack food, ramekin, ham and eggs, boiled egg, chicken provencale, rissole, pilaf, applesauce, moo goo gai pan, kedgeree, stew, tossed salad, molded salad, chicken salad, crab louis, salmagundi, potato salad, herring salad, fruit salad, tuna fish salad, tabbouleh, pasta salad, coleslaw, lamb curry, raita, fruit custard, bavarian cream, creme caramel, knish, apple turnover, pirogi, samosa

Parent: lamb curry

Description: lamb curry is curry made with lamb

Interpretation: Lamb curry is a type of dish that is made with lamb and has a curry flavor.

Reasoning: Let's think step by step in order to Reasoning: Let's think step by step in order to identify the children of lamb curry. Lamb curry is a type of dish that is made with lamb and has a curry flavor. Therefore, it is a subtype of curry and dish. Since it is a specific type of curry, it can be considered a parent concept.

Leaf: No

Children: None

---

Context:
```water soluble vitamin (Non-Leaf), vitamin c (Leaf)
b complex vitamin (Non-Leaf), biotin (Leaf)
filling (Non-Leaf), lekvar (Leaf)
beverage (Non-Leaf), rickey (Non-Leaf)
rum (Non-Leaf), grog (Leaf)```

Candidates: vitamin bc, choline, vitamin b6, biotin, niacin, vitamin b2, pantothenic acid

Parent: vitamin bc

Description: vitamin bc is a B vitamin that is essential for cell growth and reproduction

Interpretation: Vitamin bc is a type of B vitamin, which is essential for cell growth and reproduction.

Reasoning: Let's think step by step in order to Reasoning: Let's think step by step in order to identify the children of vitamin bc. Vitamin bc is a type of B vitamin, so its children should be more specific types of vitamin bc.

Leaf: No

Children: choline, vitamin b6

---

Context:
```dish (Non-Leaf), cottage pie (Leaf)
dessert (Non-Leaf), apple dumpling (Leaf)
quick bread (Non-Leaf), corn cake (Leaf)
sweet pickle (Non-Leaf), bread and butter pickle (Leaf)
conserve (Non-Leaf), apple butter (Leaf)```

Candidates: irish soda bread, oatcake, date bread, date nut bread, banana bread, nut bread, corn cake, corn dab, spoon bread, skillet corn bread, cornpone, johnnycake, ashcake, baking powder biscuit, buttermilk biscuit, rolled biscuit, drop biscuit, bran muffin, corn muffin, popover, drop scone, chicken cordon bleu, croquette, pudding, pasta, succotash, chow mein, cottage pie, spaghetti and meatballs, poi, jambalaya, roulade, swiss steak, tamale pie, bacon and eggs, enchilada, barbecue, meat loaf, patty, lobster thermidor, potpie, coquilles saint jacques, sauerbraten, coq au vin, sauerkraut, tetrazzini, moussaka, refried beans, fondue, dolmas, steak au poivre, viand, sukiyaki, timbale, porridge, scallopine, seafood newburg, lutefisk, frittata, omelet, soup, pepper steak, spanish rice, galantine, barbecued wing, salisbury steak, sashimi, couscous, fish and chips, piece de resistance, veal cordon bleu, hash, chop suey, scotch egg, scrambled eggs, poached egg, chicken marengo, casserole, terrine, macedoine, pizza, meatball, welsh rarebit, osso buco, kishke, chicken paprika, carbonnade flamande, shirred egg, scampi, mold, taco, pork and beans, bitok, french toast, burrito, scrapple, haggis, pheasant under glass, maryland chicken, beef bourguignonne, boiled dinner, rijsttaffel, chicken and rice, schnitzel, kabob, beef wellington, risotto, paella, tempura, special, souffle, mousse, fish stick, tostada, frog legs, chili, snack food, ramekin, ham and eggs, boiled egg, chicken provencale, rissole, pilaf, applesauce, moo goo gai pan, kedgeree, stew, tossed salad, molded salad, chicken salad, crab louis, salmagundi, potato salad, herring salad, fruit salad, tuna fish salad, tabbouleh, pasta salad, coleslaw, lamb curry, raita, fruit custard, bavarian cream, creme caramel, knish, apple turnover, pirogi, samosa

Parent: tamale pie

Description: tamale pie is a meat mixture covered with cornbread topping that resembles a Mexican dish

Interpretation: Tamale pie is a type of dish that has a cornbread topping, which is a type of quick bread.

Reasoning: Let's think step by step in order to Reasoning: Let's think step by step in order to identify the children of tamale pie. We can see that tamale pie is a type of dish that has a cornbread topping, which is a type of quick bread. Therefore, tamale pie is related to quick bread.

Leaf: No

Children: corn cake, corn dab, spoon bread, skillet corn bread, cornpone, johnnycake, ashcake

---

Context:
```
...
```

Candidates: vitamin, kosher, finger food, wheat germ, course, stodge, dish, dainty, puree, mince, meal, evaporated milk, yak's milk, sour milk, semi skimmed milk, colostrum, homogenized milk, coconut milk, scalded milk, chocolate milk, certified milk, pasteurized milk, raw milk, low fat milk, formula, cows' milk, nonfat dry milk, mother's milk, acidophilus milk, skim milk, corn sugar, lump sugar, caramel, granulated sugar, sugarloaf, beet sugar, brown sugar

Parent: sweetening

Description: sweetening is something added to foods to make them taste sweeter

Interpretation: Sweetening is something added to foods to make them taste sweeter, which is related to sugar and nutriment.

Reasoning: Let's think step by step in order to

\end{lstlisting}

\begin{lstlisting}[linewidth=\linewidth,language=prompt,style=generated]
// Output
Reasoning: Let's think step by step in order to identify the children of sweetening. Sweetening is something added to foods to make them taste sweeter, which is related to sugar and nutriment. Therefore, its children should be types of sweetening agents.

Leaf: No

Children: granulated sugar, lump sugar, corn sugar, caramel, brown sugar, beet sugar

\end{lstlisting}

\subsubsection{Generation: Parents}
\label{sec:gen_prompt_app_p}

\begin{lstlisting}[linewidth=\linewidth,language=prompt]
// Prompt 
Which are the most specific parent concepts of the given child concept in a taxonomy considering the context?
In your reasoning, state how the parent concepts are a supertype of the child concept.
Do not add additional comments or information, only return the output in the described format.

---

Input description.

Context: List of existing parent-child (supertype-subtype) relations in the taxonomy.

Child: Child concept (subtype) that you need to place in a taxonomy.

Description: Description of the child concept.

Taxonomy Description: Description of the taxonomy

---

Follow the following format.

Reasoning: Let's think step by step in order to ${produce the parents}. We ...

Interpretation: Description of the child concept in relation to the context taxonomy. Infer what is meant by the child concept from the context.

Parents: Comma separated list of one or more parents (supertypes) of the child concept. A parent concept must be a more general type of the child concept. If there are no suitable existing parents, invent them.

---

Context:
```baking ingredients, baking powder
baking ingredients, baking soda
baking ingredients, purpose flour
baking ingredients, unsalted butter
baking ingredients, brown sugar
cooking ingredients, spices and seasonings
cooking ingredients, oils
spices and seasonings, vanilla extract
spices and seasonings, garlic cloves
spices and seasonings, salt pepper
oils, vegetable oil
oils, olive oil```

Child: baking ingredients

Description: baking ingredients

Taxonomy Description: The topic of the taxonomy appears to be "Cooking Ingredients" or "Kitchen Pantry Items." The rough structure of the taxonomy could be organized into categories such as: 1. Oils (e.g., olive oil, vegetable oil) 2. Baking Ingredients (e.g., flour, sugar, baking powder, baking soda) 3. Spices and Seasonings (e.g., garlic cloves, salt, pepper, vanilla extract) 4. Dairy Products (e.g., unsalted butter) 5. Miscellaneous (e.g., brown sugar) This taxonomy could be further divided into subcategories or refined to include more specific ingredients or cooking-related items.

Reasoning: Let's think step by step in order to
\end{lstlisting}

\begin{lstlisting}[linewidth=\linewidth,language=prompt,style=generated]
// Output
Reasoning: Let's think step by step in order to find the most specific parent concepts of "baking ingredients". We can observe that "baking ingredients" is already a category in the given taxonomy, and it has several subtypes such as "baking powder", "baking soda", "purpose flour", "unsalted butter", and "brown sugar". This suggests that "baking ingredients" is a subtype of a more general category. Since the taxonomy is about "Cooking Ingredients" or "Kitchen Pantry Items", a suitable parent concept for "baking ingredients" could be "cooking ingredients".

Interpretation: The child concept "baking ingredients" refers to a category of ingredients used in baking, which is a part of the broader topic of cooking ingredients.

Parents: cooking ingredients

\end{lstlisting}

\subsubsection{Generation: Children}
\label{sec:gen_prompt_app_c}
\begin{lstlisting}[linewidth=\linewidth,language=prompt]
// Prompt
Which of the candidates are child concepts (subtypes) of the given parent concept (supertype) in a taxonomy?
The context shows existing parents and children concepts and whether the children are leaves.
In your reasoning, state how the parent concept is a supertype of the selected child concepts.
Do not add additional comments or information, only return the output in the described format.

---

Input description.

Context: List of existing parent-child (supertype-subtype) relations in the taxonomy.

Candidates: Candidate children of the concept separated by commas to select from.

Parent: Parent concept that you need to place in a taxonomy.

Description: Description of the parent concept.

Interpretation: Description of the child concept in relation to the taxonomy.

Previous Reasoning: past Reasoning: with errors

Previous Leaf: past Leaf: with errors

Previous Children: past Children: with errors

Instructions: Some instructions you must satisfy

---

Follow the following format.

Reasoning: Let's think step by step in order to ${produce the children}. We ...

Leaf: Whether the parent concept should be added as a leaf (has no children). Answer with Yes or No.

Children: Comma separated list of candidates that are children of the parent concept in a taxonomy.A child concept must be a type of the parent concept.Separate with commas.

---

Context:
```baking ingredients (Non-Leaf), baking powder (Leaf)
baking ingredients (Non-Leaf), baking soda (Leaf)
baking ingredients (Non-Leaf), purpose flour (Leaf)
baking ingredients (Non-Leaf), unsalted butter (Leaf)
baking ingredients (Non-Leaf), brown sugar (Leaf)
cooking ingredients (Non-Leaf), spices and seasonings (Non-Leaf)
cooking ingredients (Non-Leaf), oils (Non-Leaf)
spices and seasonings (Non-Leaf), vanilla extract (Leaf)
spices and seasonings (Non-Leaf), garlic cloves (Leaf)
spices and seasonings (Non-Leaf), salt pepper (Leaf)
oils (Non-Leaf), vegetable oil (Leaf)
oils (Non-Leaf), olive oil (Leaf)```

Candidates: spices and seasonings, oils

Parent: baking ingredients

Description: baking ingredients

Interpretation: The child concept "baking ingredients" refers to a category of ingredients used in baking, which is a part of the broader topic of cooking ingredients.

Previous Reasoning: Reasoning: Let's think step by step in order to determine the children of "baking ingredients". We can see that "baking ingredients" is a category of ingredients used in baking, which is a part of the broader topic of cooking ingredients. Since "spices and seasonings" and "oils" are both categories of ingredients, but not specifically used only in baking, they cannot be considered as children of "baking ingredients".

Previous Leaf: No

Previous Children: Reasoning: Let's think step by step in order to determine the children of "baking ingredients". We can see that "baking ingredients" is a category of ingredients used in baking, which is a part of the broader topic of cooking ingredients. Since "spices and seasonings" and "oils" are both categories of ingredients, but not specifically used only in baking, they cannot be considered as children of "baking ingredients". Leaf: No Children:

Instructions: reasoning lets think step by step in order to determine the children of baking ingredients. we can see that baking ingredients is a category of ingredients used in baking, which is a part of the broader topic of cooking ingredients. since spices and seasonings and oils are both categories of ingredients, they cannot be considered as children of baking ingredients., but not specifically used only in baking are not valid children, since they are not in the candidates.

Reasoning: Let's think step by step in order to
\end{lstlisting}
\begin{lstlisting}[linewidth=\linewidth,language=prompt,style=generated]
// Output
Reasoning: Let's think step by step in order to determine the children of "baking ingredients". We can see that "baking ingredients" is a category of ingredients used in baking, which is a part of the broader topic of cooking ingredients. Since "spices and seasonings" and "oils" are categories of ingredients used in cooking in general, not specifically in baking, they cannot be considered as children of "baking ingredients".

Leaf: No

Children:
\end{lstlisting}

\end{document}